\documentclass{article}

\usepackage{PRIMEarxiv}

\usepackage[utf8]{inputenc} 
\usepackage[T1]{fontenc}    
\usepackage{hyperref}       
\usepackage{url}            
\usepackage{booktabs}       
\usepackage{amsfonts}       
\usepackage{nicefrac}       
\usepackage{microtype}      
\usepackage{lipsum}
\usepackage{fancyhdr}       
\usepackage{graphicx}       
\usepackage{amssymb}
\usepackage{array}
\usepackage{graphicx}%
\usepackage{mathtools}
\usepackage{tikz}
\usepackage{multirow}%
\usepackage{amsmath,amssymb,amsfonts}%
\usepackage{amsthm}%
\usepackage{mathrsfs}%
\usepackage[title]{appendix}%
\usepackage{xcolor}%
\usepackage{textcomp}%
\usepackage{manyfoot}%
\usepackage{booktabs}%
\usepackage{algorithm}%
\usepackage{algorithmicx}%
\usepackage{algpseudocode}%
\usepackage{listings}%
\usepackage{subcaption}
\usepackage{todonotes}
\usepackage{url}
\graphicspath{{media/}}     

\newcommand{\nuria}[1]{\textcolor{black}{#1}}

\newcommand{\context}[1]{\textcolor{black}{#1}}
\newcommand{\challenge}[1]{\textcolor{black}{#1}}
\newcommand{\proposal}[1]{\textcolor{black}{#1}}
\newcommand{\evaluation}[1]{\textcolor{black}{#1}}
\newcommand{\paperdesc}[1]{\textcolor{black}{#1}}

\newcommand{\ecml}[1]{\textcolor{black}{#1}}
\newcommand{\miner}[1]{\textcolor{black}{#1}}
\newcommand{\revista}[1]{\textcolor{black}{#1}}
\newcommand{\paco}[1]{\textcolor{black}{#1}}

\title{Krum Federated Chain (KFC): Using blockchain to defend against adversarial attacks in Federated Learning}

\author{
  Mario García-Márquez, Nuria Rodríguez-Barroso \\
  Department of Computer Science and Artificial Intelligence,\\ Andalusian Research Institute in Data Science \\
  and Computational Intelligence (DaSCI) \\
  University of Granada \\
  Granada\\
  \texttt{\{mariogmarq, rbnuria\}@ugr.es} \\
   \And
  M.V. Luzón \\
    Department of Software Engineering,\\ 
    Andalusian Research Institute in Data Science \\
  and Computational Intelligence (DaSCI) \\
  University of Granada \\
  Granada\\
  \texttt{luzon@ugr.es} \\
  \And
  Francisco Herrera \\
  Department of Computer Science and Artificial Intelligence,\\ Andalusian Research Institute in Data Science \\
  and Computational Intelligence (DaSCI) \\
  University of Granada \\
  Granada\\
  \texttt{herrera@decsai.ugr.es} \\
}

\begin{document}
\maketitle

\begin{abstract}
Federated Learning presents a nascent approach to machine learning, enabling collaborative model training across decentralized devices while safeguarding data privacy. However, its distributed nature renders it susceptible to adversarial attacks. Integrating blockchain technology with Federated Learning offers a promising avenue to enhance security and integrity. In this paper, we \paco{tackle} the potential of blockchain in defending Federated Learning against adversarial attacks. \ecml{First}, we test Proof of Federated Learning, a well known consensus mechanism designed ad-hoc to federated contexts, as a defense mechanism demonstrating its efficacy against \ecml{Byzantine} and backdoor attacks when at least one miner remains uncompromised. \ecml{Second}, we propose Krum Federated Chain, a novel defense strategy combining Krum and Proof of Federated Learning, \paco{valid to} defend against any configuration of \ecml{Byzantine} or backdoor attacks, even when all miners are compromised. Our experiments conducted on image classification datasets validate the effectiveness of our proposed approaches.
\end{abstract}


\section{Introduction}

\context{In the realm of artificial intelligence, Federated Learning (FL) \cite{kairouz2021advances} stands as a promising paradigm revolutionizing traditional machine learning approaches. Its distributed nature allows for training models collaboratively across decentralized devices while preserving data privacy. Furthermore, the integration of blockchain \cite{di2017blockchain} technology into FL has emerged as a promising solution to enhance its security and integrity \cite{nguyen2021federated}. By leveraging the immutable and transparent nature of blockchain, FL systems can establish a tamper-resistant record of model updates and participant contributions. This not only enhances the trustworthiness \cite{yang2022trustworthy, Sabater2022} of the collaborative learning process but also mitigates the risk of data manipulation and unauthorized access. Additionally, blockchain facilitates the establishment of smart contracts, enabling automatic verification and enforcement of agreements between participating nodes. Such advancements not only bolster the security of FL but also streamline its operation, paving the way for widespread adoption across diverse industries.}

\challenge{Despite FL's significant benefits, such as data privacy by processing data locally on user devices, this same decentralized feature can also be a weakness \cite{rodriguez2023survey}. In a typical adversarial attack, a malicious actor could manipulate the data or learning models at one of the system's nodes without other nodes being aware of the manipulation. Since the data isn't centrally collected, detecting these anomalies can be more challenging, making FL systems particularly vulnerable to such attacks \cite{wang2020attack}. Understanding and addressing these vulnerabilities are crucial to ensuring the security and robustness of FL systems.}


As stated before, blockchain technology~\cite{duc-2023} has been extensively utilized in combination with FL, to enhance data integrity and transparency across distributed networks. However, its use has not been significantly explored as a defense against adversarial attacks in FL systems. 

We hypothesize that the immutable and transparent nature of blockchain could potentially be harnessed to develop mechanisms that detect and prevent such attacks. By leveraging blockchain's decentralized ledger, it might be possible to create a more secure environment where alterations to data or models are immediately noticeable and traceable, thereby providing a robust layer of security against adversarial threats in FL applications. \ecml{First}, we test Proof of Federated Learning (PoFL) \cite{qu-2021}, a well known consensus mechanism designed ad-hoc to FL \revista{created with energy efficiency in mind}, as a defense against \ecml{Byzantine} and backdoor attacks \cite{bagdasaryan2020backdoor}. We find that PoFL is useful when there is at least one miner not being attacked, but the federated scheme remains vulnerable when all the miners are being attacked.

\proposal{\ecml{Second,} \paco{the proposal to tackle the} vulnerabilities of PoFL identified regarding the configuration of the attack, Krum Federated Chain (KFC). \paco{It is} a defense based on the combination of the Krum aggregation operator \cite{blanchard-2017} and blockchain with PoFL consensus mechanism, to defend FL against any configuration of \ecml{Byzantine} and backdoor attacks, including the configuration in which all the miners are being attacked.}


\evaluation{We test our contribution \paco{using two different attacks, over three image classification datasets}, namely: EMNIST \cite{emnist}, Fashion MNIST \cite{fashionmnist-2017}, and CIFAR-10~\cite{torralba-2008}. As adversarial attacks we consider:
\begin{itemize}
    \item Byzantine attacks \cite{blanchard-2017, fang2020}, which aim to degrade model performance by sending random updates. \paco{Specifically, we consider \ecml{label-flipping Byzantine attacks~\cite{fang2020} which involve randomly altering the labels of data points.}}
    \item  Backdoor attacks \cite{bagdasaryan2020backdoor}, which consists of injecting a secondary (or backdoor) task while maintaining the performance of the model in the original task. \paco{We focus on pattern-key backdoor attacks} \cite{bagdasaryan2020backdoor} which are based on injecting the secondary task using specific patterns out of the input distribution.
\end{itemize}
  Regarding the attack scenarios, \paco{we consider two}: 
 \begin{itemize}
 \item Just one miner being attacked.
 \item All the miners being attacked.
 \end{itemize}
  As baselines, we use other consensus mechanisms, such as: Proof of Work and Proof of Stake, raw FL without blockchain (namely, client-server) \ecml{and furthermore, we evaluate prominent defense mechanisms commonly utilized in FL, namely: Krum~\cite{blanchard-2017} and Trimmed-mean~\cite{trimmedmean}.}}

\paperdesc{\paco{T}he paper is organized as follows: in Section \ref{background} we introduce some background concepts including FL in Section \ref{sec:fl}, \ecml{adversarial} attacks in FL in Section \ref{sec:adversarial_fl} and related works about blockchain applied to FL in \ref{sec:blockfl}. In Section \ref{sec:pofl} we explain the hypothesis of using PoFL as a defense mechanism. In Section \ref{sec:kfc} we detail the proposed defense mechanism KFC. Hereafter, we specify the experimental set up in which we test our contributions in Section \ref{experimental_setup} including: evaluation datasets (see Section \ref{sec:evaldatasets}), \ecml{label-flipping Byzantine attacks (see Section \ref{sec:byzantineattacks})}, pattern-key backdoor attacks (see Section \ref{sec:backdoorattacks}), attack scenarios  (see Section \ref{sec:scenarios}), evaluation metrics (see Section \ref{sec:metrics}), baselines (see Section \ref{sec:baselines}) and implementations details (see Section \ref{sec:details_imp}). Finally, we test the hypothesis of PoFL in Section \ref{sec:analysis_pofl} and prove the proposal KFC in Section \ref{sec:analysis_kfc}. \ecml{The present study's limitations are discussed in detail in Section \ref{sec:limitations}.} The conclusions and future work are detailed in Section \ref{sec:conclusion}}.


\vspace{-0.8em} 
\section{Background}\label{background}
In this section we provide the necessary concepts to follow the rest of the paper. In Section~\ref{sec:fl} we provide the basic notions about FL, \revista{Section~\ref{sec:appblockchain} provides an introduction to the fundamental concepts of blockchain technology}, in Section~\ref{sec:adversarial_fl} we define the \ecml{adversarial} attacks in the context of FL \ecml{subsequently delving into untargeted attacks and backdoor attacks in Sections~\ref{sec:byzantine_fl} and~\ref{sec:backdoor_fl}} respectively, \revista{followed by an explanation of the model replacement technique in Section~\ref{sec:appreplacement}} and, finally, in Section~\ref{sec:blockfl} we relate the concepts of FL and blockchain according to recent related works.

\subsection{Federated Learning}\label{sec:fl}

FL embodies a distributed machine learning paradigm aimed at constructing machine learning models without the direct exchange of training data among participating entities \cite{yang19}. It operates within a network of data proprietors \miner{called clients}, \ecml{denoted by $\{ C_1, \ldots, C_n \}$ with their respective datasets $\{ D_1, \ldots, D_n\}$}, engaging in two primary phases:

\begin{enumerate}
\item \textit{Model training phase:} During this phase, each client collaborates by sharing information without revealing their raw data, thereby collectively training a machine learning model. This model may be housed at a single client or distributed across multiple clients.

\ecml{For this, each client, $C_i$, trains a local model, $L_i$, \paco{parametrized by a vector $V_i$}, using its own data $D_i$. These local model \paco{parameters} are then aggregated to form a global model \paco{$G$, parametrized by $V_G = \Delta(V_1, \ldots, V_n)$,} using a fixed federated aggregation operator, $\Delta$. The global model \paco{parameters} is distributed back to the clients, and the process is repeated iteratively until a predefined stopping criterion is met.}

\item \textit{Inference phase:} Subsequently, clients work collaboratively to apply the jointly trained model, to process new data instances.
\end{enumerate}

\begin{figure}[h!]
    \centering
    \includegraphics[width=1.2\textwidth, trim={0 5cm 0 5cm}, clip]{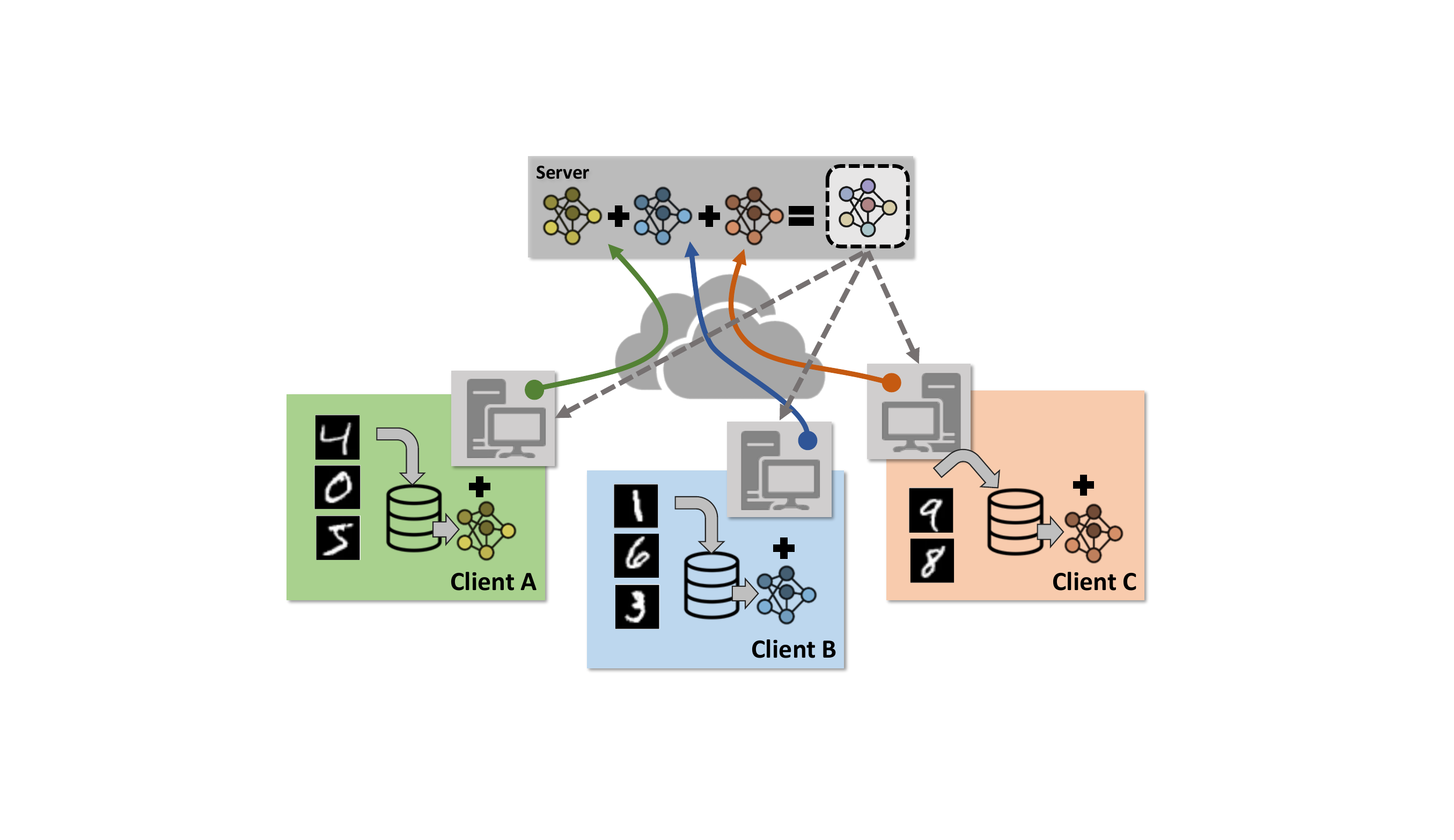}
    \caption{Illustration of a typical FL workflow. Figure inspired by \cite{tutorialnuria}.}
    \label{fig:enter-label}
\end{figure}

Both phases can operate synchronously or asynchronously, contingent upon factors such as data availability and the status of the trained model.

It's pivotal to note that while privacy preservation is central to this paradigm, another crucial aspect involves establishing a equitable mechanism for distributing the profits accrued from the collaboratively trained model.

\subsection{Blockchain technology}\label{sec:appblockchain}
A blockchain is a distributed ledger technology that stores a sequence of blocks containing a set of verifiable transactions. The blockchain is replicated and shared across multiple nodes \miner{called miners} in a peer-to-peer network~\cite{jiang-2018, duc-2023}, ensuring that every node maintains a copy of the blockchain. Formally, a blockchain can be defined as a finite sequence of data blocks $b_1, b_2, b_3, \ldots, b_n$, each possessing the following attributes:
\begin{itemize}
    \item \emph{Block ID}: Denoted by $b_i.id$, it contains the hash value of \ecml{some part of} the block's content, generated through a predefined and publicly known cryptographic hash function $H$. \ecml{A prevalent approach utilizes the block header to efficiently derive a unique identifier for the block. This identifier typically incorporates the Merkle root hash of the data contained within the block. Formally, this identifier can be represented as $b_i.id = H(b_i)$, where $H$ denotes a cryptographic hash function and $b_i$ signifies the i-th block. Notably, due to the readily computable nature of this identifier, it is often omitted from the stored block data.}
    
    \item  \emph{Previous hash}: Denoted by $b_i.prev$, it contains the identifier of the preceding block, $b_{i-1}$, to which $b_i$ is appended. This can be expressed as $b_i.prev = b_{i-1}.id$.
    \item \emph{Transaction data}: This attribute encompasses the data intended for storage within the block, the structure of which is contingent upon the specific blockchain implementation. \ecml{However, a common approach involves utilizing a Merkle tree to organize and verify transactions efficiently.} The transaction data must be verifiable through a predetermined validation process, ensuring the integrity and authenticity of the stored information. In some instances, the block may incorporate the entirety of the blockchain's state, whereas in others, it may solely contain transactions, enabling users to derive the blockchain's state by inspecting the entire ledger enabling traceability and transparency of the data.
\end{itemize}

The incorporation of previous hash information enables the preservation of data integrity within the blockchain~\cite{bitcoin}. Specifically, if a modification is made to a block $b_j$ subsequent to its appending to the blockchain, a discrepancy will arise between $b_{j+1}.prev$ and $b_j.id$. In such an event, the blockchain will be deemed invalid, and no further operations will be allowed. Notably, the decentralized nature of the blockchain, wherein it is replicated across multiple nodes on a peer-to-peer network, facilitates the retrieval of a valid version of the blockchain from another node in the event of an invalidation, thereby ensuring the maintenance of data integrity and immutability.

The decentralized and asynchronous nature of blockchain systems, absent of a central authority, presents a significant challenge in maintaining a consistent state across the network and ensuring reliable communication, particularly in regards to determining which block to append to the ledger. To address this issue, blockchain systems employ a consensus mechanism~\cite{xiao-2020}, e.g., Proof of Work (PoW)~\cite{bitcoin} and Proof of Stake (PoS)~\cite{stake}, which typically combines cryptographic techniques with economic incentives. By leveraging these consensus mechanisms, blockchain technology enables the creation of decentralized and trustless databases that ensure reliability and integrity. As a result, blockchain has been widely adopted in various fields, such as finance and supply chain management, where it helps establish trust between different parties.

\vspace{-0.8em}
\subsection{\ecml{Adversarial attacks in Federated Learning}}\label{sec:adversarial_fl}
\ecml{Adversarial attacks pose a significant challenge to FL. The decentralized nature of FL can be exploited by malicious clients who may send poisoned updates to deliberately modify the global model. While adversarial attacks is a general problem in artificial intelligence~\cite{Heinrich2024}, in FL the server's inability to directly inspect the client-side training data worsen this vulnerability. These attacks typically involve clients with white-box access to the aggregated model, their locally trained models, and their datasets. In many cases, attackers are assumed to possess even more information, including knowledge of the aggregation mechanism used by the server. These attacks can be categorized based on their objectives into targeted attacks, also known as backdoor attacks, and untargeted attacks.}

In order to ensure the effectiveness of the attack and the no mitigation of the attack in the aggregation with benign clients' \ecml{adversarial} attacks are usually combined with model replacement techniques~\cite{bagdasaryan2020backdoor}.

\subsubsection{\ecml{Untargeted attacks in Federated Learning}}\label{sec:byzantine_fl}
\ecml{The primary objective of untargeted attacks is to degrade the overall performance of the global model on its intended task. A particularly severe type of untargeted attack is known as a Byzantine attack, where malicious clients share randomly generated model updates or train on randomly modified data. These attacks can be detected by analyzing the performance of local model updates on the server. However, differentiating between Byzantine attacks and clients with very particular training data distributions can be challenging.~\cite{survey-nuria}}

\ecml{Formally, in the context of FL, the primary objective is to collaboratively train a global model $G$ by aggregating local models $L_i$ from multiple clients. This global model aims to approximate a target function $m: \mathcal{X} \rightarrow \mathcal{Y}$, where $\mathcal{X}$ represents the input space and $\mathcal{Y}$ denotes the output space. An untargeted attack in FL seeks to maliciously perturb a set of the local models such that the resulting global model $\hat{G}$ approximates instead a function $\hat{m}$ characterized by $\hat{m}(x) \ne m(x)$ for all inputs $x \in \mathcal{X}$.}

\subsubsection{Backdoor attacks in Federated Learning}\label{sec:backdoor_fl}


Backdoor attacks \cite{bagdasaryan2020backdoor} consist of injecting a secondary or backdoor task while maintaining the performance in the original task. Their design has a wide range of options depending on the injected (backdoor) task \cite{gong2022backdoor}. \ecml{In \cite{survey-nuria} a taxonomy for backdoor attacks is presented, categorizing them into two primary types: (1) input-instance-key, where the model target specific input instances, manipulating their labels to a predetermined target label different from the original, and (2) pattern-key, where the model associate a particular pattern within an input sample with a specific target label.}

\ecml{Formally, the goal is to create a model $\hat{G}$ that correctly approximates the desired function~\cite{baddasaryan2020}, $m$  while simultaneously learning a hidden function $m^*: \mathcal{X^*} \to \mathcal{Y}$, where $\mathcal{X^*}$ is the domain of inputs that triggers the backdoor. The trigger patterns can be part of the original input space $\mathcal{X}$ in the case of input-instance-key strategies or a modified version of it in the case of pattern-key strategies. By learning both $m$ and $m^*$, the model $\hat{G}$ will try to learn the function $\hat{m}$ defined by Equation \ref{eq:backdoor_obj}.}

\begin{equation}\label{eq:backdoor_obj}
    \ecml{\hat{m}(x) = \begin{cases*}
        m^*(x) & if $x \in \mathcal{X}^*$ \\
        m(x) & if $x \in \mathcal{X}\setminus{\mathcal{X}^*}$
    \end{cases*}}
\end{equation}

\subsubsection{Model replacement technique for attacks in Federated Learning}\label{sec:appreplacement}
The primary objective of this technique is to amplify the influence of the adversarial attack to avoid mitigation effects by other clients' updates. Let $G^t$ and $L^t_i$ be the global model and local model respectively, of the $i$-th client at a given round $t$, $n$ the number of clients participating in the round and $\eta$ the server learning rate. Then, the update of the global model \paco{parameters} in the round $t$ is performed as in Eq. \ref{eq:fedavg}:
\begin{equation}\label{eq:fedavg}
    \paco{V_G^{t+1}={V_G}^{t}+ \frac{\eta}{n}\sum_{i=1}^n({V}_i^t - {V_G}^t)}
\end{equation}

We consider that, in the learning round $t$, only one adversarial client is selected which will try to replace the global model $G^t$ with his \ecml{adversarial} model $L^t_{adv}$, which is optimized \ecml{for the goal of the attack}. To do so, a new model is computed as follows
\begin{equation}\label{eq:boosting}
     \hat{\paco{V}}_{adv}^t = \beta(\paco{V}_{adv}^t - \paco{V_G}^t)
\end{equation}

where $\beta = \frac{n}{\eta}$ is the boost factor. Replacing Eq. \ref{eq:boosting} in Eq. \ref{eq:fedavg} we deduce
\begin{equation}\label{eq:replacementexp}
    \paco{V_G}^t = \paco{V_G}^{t-1} + \frac{\eta}{n} \frac{n}{\eta}(V^t_{adv} - \paco{V_G}^{t-1}) + \frac{\eta}{n} \sum_{i=2}^n(\paco{V_i}^t - \paco{V_G}^{t-1})
\end{equation}

Now, if we assume that the model converges, eventually we may also assume that $V_i^t - V_G^{t-1} \approx 0$ for benign clients. Thus, we can approximate Eq. \ref{eq:replacementexp} by
\begin{equation}\label{eq:replacement}
    \paco{V_G}^t \approx \paco{V_G}^{t-1} + V^t_{adv} - \paco{V_G}^{t-1} = V^t_{adv}
\end{equation}

which replaces the global model with the model trained by the adversarial client. In the case where multiple adversarial clients participate, the boosting factor $\beta$ is divided by the number of attackers. In this paper, we consider $\eta = 1$.

\vspace{-0.8em}
\subsection{Related works on blockchain applied to federated learning}\label{sec:blockfl}

The characteristics of blockchain have derived into proposals about how to implement blockchain in a \nuria{FL} system~\cite{kang-2024, nguyen-2021, zhu-2023}. This previous works tackle problems present in classical \nuria{FL} such as an environment with the server as a single point of failure or poor scalability in some scenarios~\cite{survey-nuria}. \ecml{As outlined in \cite{zhu-2023}, a taxonomy based on network topology classifies blockchain-federated learning architectures into three primary categories: (1) decoupled, where nodes are restricted to operating exclusively within either the blockchain or the FL system, (2) coupled, all nodes participate in both the blockchain and FL systems, and (3) overlapped, which encompasses nodes that can operate within the blockchain, the FL system, or both. It is noteworthy to observe the predominance of decoupled architectures in the literature~\cite{lu2020, kumar2021, pokhrel2020}}. \miner{The nodes operating within the blockchain are called miners, meanwhile the ones operating within the FL system are called clients. Thus according to the architecture this terms may be (in the case of a coupled architecture) or not be (in the case of a decoupled or overlapped architecture) equivalent.}

A relevant number of this proposed architectures are found to use PoW as a consensus mechanism, which may be considered unfeasible due to the computational cost of a network using it~\cite{mora2018}. Thus, is common to see consensus \nuria{mechanisms} with a higher energy efficiency such as PoS to be utilized instead, but they usually require the presence of a economic reward mechanism which may not always be assumed. This have led to the creation of consensus mechanisms specific to \nuria{FL}, where \nuria{PoFL \cite{qu-2021} is the most widely used}. 

Consensus mechanism can be considered \ecml{one of the most} important parts of a blockchain architecture applied to FL since it usually determines the way training and aggregation is performed during the learning process of the federated model and thus, \ecml{a significant part of the} architecture of the network. However, there is little literature about the impact of this part of a blockchain network on adversarial attacks against the federated model.


\vspace{-0.8em}
\section{PoFL as a defense against \ecml{adversarial} attacks\paco{: A testing hypotesis}}\label{sec:pofl}

In this section we test the hypothesis that PoFL, the consensus mechanism, can be a defense against adversarial attacks in FL.





Inspired by the Proof-of-Useful Work concept~\cite{Sabry2023}, PoFL retains the core principle of PoW by requiring participants to solve computationally intensive problems to achieve consensus. However, unlike PoW, which utilizes computational power for tasks with no inherent value (e.g., finding a specific hash nonce in Bitcoin \cite{duc-2023}) \revista{and thus being usually discarded for real applications}, PoFL redirects these resources towards training a \nuria{FL} model \revista{empowering energy efficiency}.

\begin{figure}[ht]
    \centering
    \includegraphics[width=0.70\textwidth]{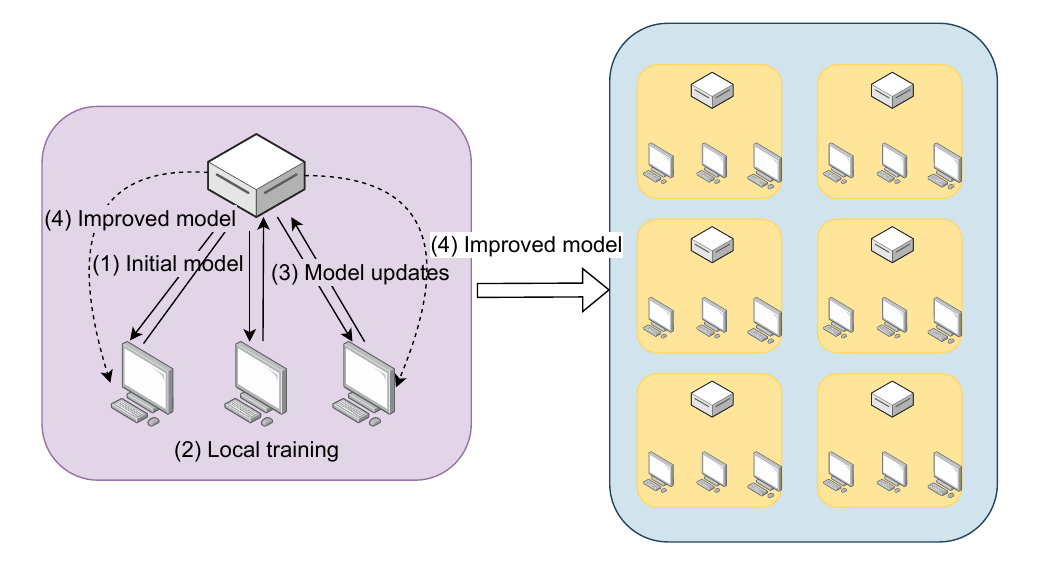}
    \caption{Diagram \ecml{illustrating} the pooled-mining \ecml{mechanism}.}
    \label{fig:pooledmining}
\end{figure}

It is important to acknowledge that PoFL leverages a \miner{decoupled} pooled-mining architecture. Within this structure, network \miner{clients} are divided into distinct pools, each overseen by a designated miner. These pools operate independently, training their respective federated models without inter-pool communication. The pool that trains the model exhibiting the highest accuracy on a predetermined test dataset emerges victorious in the consensus round. Consequently, the winning model is integrated into the blockchain and broadcasted throughout the network to all clients. The process in each pool is showed in \nuria{Figure} \ref{fig:pooledmining}:
\begin{enumerate}
    \item The miner broadcasts an initial model to the clients.
    \item Clients train the locally held copy of the model using their private data. This training step involves calculating the local updates based on the difference between the received model and the trained one.
    \item The clients send the model updates to the miner. The miner aggregates these updates to create a new, improved model. The miner evaluates the performance of the aggregated model against a predetermined accuracy goal. If the goal is achieved, consensus is reached for that round.
    \item Upon reaching consensus, the miner appends the improved model to the blockchain and broadcasts it to all miners \revista{and to its corresponding pool}, and thus all participating clients\revista{, as shown in the step 4 of Figure \ref{fig:pooledmining}}. If consensus is not achieved (i.e., accuracy goal not met), the process restarts from step 2, with clients training until the desired accuracy is attained.
\end{enumerate}

Formally, we can model a blockchain network using PoFL as a set of pools $P_1, P_2, \ldots, P_n$, where each pool $P_i$ contains a miner $m_i$ and a set of clients $C_1^i, \ldots, C_{j_i}^i$. In every learning round $t$, every pool $P_i$ will compute a federated model denoted by $L_i^t$ through a federated training process coordinated by the miner $m_i$. Then every miner $m_i$ computes the accuracy of its model $L_i^t$ on a determined dataset $\mathcal{D}$, which we denote by $accuracy(L_i^t, \mathcal{D})$. Then, the miner $m_{i*}$ wins consensus if 
\begin{equation*}
    accuracy(L_{i*}^t, \mathcal{D}) \ge accuracy(L_i^t, \mathcal{D}) \quad \forall i \in \{ 1, 2, \ldots, n\}.
\end{equation*}

Upon reaching consensus, the model $L_{i*}^t$ is broadcasted to all miners $m_j$ with $j \ne i*$ and check that the claimed accuracy is true. After a successful verification, the miner $m_{i*}$ creates a new block $b_{t}$ which sets $G^{t+1}=L_i^t$ and appends it to the ledger. Finally, all the miners download the new version of the blockchain and broadcast the new global model $G^{t+1}$ to their clients, ensuring that all clients have access to the latest model.

The PoFL consensus mechanism exhibits potential resistance against \ecml{adversarial} attacks. Backdoor attacks involve an adversary within a pool attempting to manipulate the training process of the federated model. This can lead to a slight degradation in the model's performance on its original task. By leveraging a performance-based consensus mechanism and pooled-mining, PoFL inherently discourages such attacks. Pools containing adversarial clients are likely to produce models with lower accuracy on the predetermined test dataset compared to pools without malicious clients. \ecml{This is evident in the case of untargeted attacks, where the sole objective is to degrade the overall model's performance. Alternatively, a backdoor attack can be conceptualized as a specific instance of multi-task learning with hard parameter sharing~\cite{baddasaryan2020}. In this context, the primary task and the secondary, malicious task may be conflicting, potentially leading to degraded performance on the primary task~\cite{yu2020}.} Consequently, the PoFL consensus mechanism would likely favor the pool with the superior performing model, effectively filtering out the \ecml{adversarial model in both cases} and mitigating the attack.

\vspace{-0.8em}
\section{Krum Federated Chain \paco{Defense Strategy}}\label{sec:kfc}
It is crucial to acknowledge that the potential resistance of PoFL against \ecml{adversarial} attacks hinges on the optimistic assumption that a pool lacking of adversarial clients consistently exists. This assumption can be considered unrealistic in real-world scenarios. Therefore, we propose the KFC \paco{defense strategy} as a more robust security alternative to PoFL. While KFC maintains the same architectural foundation and energy efficiency principles as PoFL, it incorporates additional security mechanisms to mitigate adversarial attacks even in the presence of malicious actors within all the pools in the network.

KFC leverages the Krum aggregation operator~\cite{blanchard-2017}, to enhance its resilience against attacks on the federated model. This operator functions by sorting client updates based on the geometric distances between their respective model update distributions. Subsequently, it selects the update closest to the majority, effectively filtering out outliers. Adversarial clients attempting to manipulate the model are likely to generate updates that deviate significantly from the norm, making them susceptible to identification and exclusion by the Krum operator.


Formally, we represent the update received from the $i$-th client as a vector $V_i$ inside a finite dimensional real vector space $\mathbb{R}^d$. The Krum aggregation rule, denoted by $KR(V_1, \ldots, V_n)$, is then defined as follows. For any two clients $i$ and $j$, where $i \ne j$, we use the notation $i \to j$ to indicate that the vector $V_j$ belongs to the set of $n-f-2$ closest vectors to $V_i$ in terms of euclidean distance. Building upon this concept, a score can be assigned to each client:
\begin{equation*}
    s(i) = \sum_{i\to j} || V_i - V_j ||^2.
\end{equation*}
That is, the sum of the square of distances between the $n-f-2$ closest vectors to $V_i$. Then, $KR(V_1, \ldots, V_n)$ is set to be $V_{i_*}$ where $i_*$ is the index of the minimal score, that is, $s(i_*) \le s(i)$ for every $i$. Note that $f$ is a parameter that we must take into account when using the Krum operator, in this paper we default it to be 1. \ecml{This parameter serves as a threshold for the expected volume of adversarial clients. In practice, a conservative static value is typically employed.}

\revista{As stated in Krum original paper~\cite{blanchard-2017}, the Krum operator exhibits a time complexity of $O(n^2 \cdot (d + \log n))$ where $d$ denotes the data dimensionality and $n$ represents the number of clients. Notably, in the KFC framework, $n$ specifically refers to the client count within an individual pool, rather than the entire network, thereby facilitating horizontal scalability of the system.}

KFC inherits the pooled-mining architecture from PoFL while incorporating the Krum aggregation operator. This design choice allows KFC to benefit from the inherent resistance to adversarial attacks offered by the underlying blockchain architecture and consensus mechanism. Moreover, the Krum operator reinforces KFC's security posture by filtering out outliers within each pool's client update distributions. This mechanism alleviates the need for the assumption of a pool entirely devoid of malicious actors. This combined approach leverages multiple defense mechanisms, resulting in a more robust solution for \nuria{FL} under adversarial conditions.

\vspace{-0.8em}
\section{Experimental set up}\label{experimental_setup}

\nuria{The evaluation of the hypothesis of \nuria{PoFL as defense mechanism and} the proposed KFC is performed by means of the accuracy of the resulting FL model in three \nuria{image classification} datasets arranged for FL. This section specify: the datasets employed in Section \ref{sec:evaldatasets}, \ecml{the Byzantine attack used in Section \ref{sec:byzantineattacks}}, \ecml{the backdoor attack used} in Section \ref{sec:backdoorattacks}, the details about attacks scenarios in Section \ref{sec:scenarios}, the evaluation metrics employed in Section \ref{sec:metrics}, the baselines which we compare with in Section \ref{sec:baselines} and, finally, the implementations details in Section \ref{sec:details_imp}.}



\vspace{-0.8em}
\subsection{Evaluation datasets}\label{sec:evaldatasets}

\nuria{For the evaluation of both analyses we use classical image classification datasets.} Since PoFL and KFC need a validation set for computing the accuracy of each model, \nuria{we fix the 20\% of the test data to validation data and the remaining 80\% to the test data}. The three datasets used in the evaluation are described as what follows:

\begin{enumerate}
    \item The EMNIST dataset, which contains a balanced subset of the digits dataset containing 28,000 samples of each digit. The dataset consists of 280,000 samples, which 240,000 of them are used for training and 40,000 as test samples (\nuria{we use 8,000 as validation samples and the remaining 32,000 as test samples}). We set the number of clients to 200 proportionally distributed among 3 miners.
    
    \item The Fashion MNIST \cite{fashionmnist-2017} which contains a balanced subset of the 10 different classes containing 7,000 samples of each class. Thus, the dataset consists of 70,000 samples, which 60,000 are training samples and 10,000 test samples (\nuria{we use 2,000 as validation samples and 8,000 as test samples}). We set the number of clients to 200 proportionally distributed among 3 miners.
    
    \item The CIFAR-10 dataset which consists of 60,000 color images in 10 classes, with 6,000 images per class. 50,000 of the samples are used as training images and 10,000 as test images (\nuria{we use 2,000 as validation samples and 8,000 as test samples}), which correspond to 1,000 of each class. We set the number of clients to 100 (in order to have more data per client as the problem is more challenging) proportionally distributed among 2 miners.
        
\end{enumerate}

\vspace{-0.8em}
\subsection{\ecml{Label-Flipping Byzantine Attacks}}\label{sec:byzantineattacks}
\begin{figure}[h]
    \centering
    \includegraphics[width=0.5\linewidth]{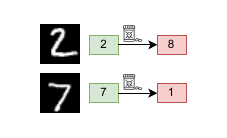}
    \caption{\ecml{Procedure of a label-flipping attack on a machine learning model.}}
    \label{fig:labelflip}
\end{figure}

\ecml{Regarding the untargeted attack, we implemented a Byzantine attack based on label-flipping. In this attack, the labels of randomly selected training data points were intentionally modified to a label different from the original. This deliberate mislabeling would induce adversarial clients to generate random model updates, ultimately leading to a degradation in the performance of the global model. The general procedure of this attack is illustrated in Figure \ref{fig:labelflip}.}

\subsection{Pattern-key Backdoor Attacks}\label{sec:backdoorattacks}

\begin{figure}[h]
    \centering
    \resizebox{0.6\textwidth}{!}{
    \begin{tabular}{m{3.0cm}m{3.0cm}m{3.0cm}m{0.4cm}m{0.5cm}}

    \includegraphics[width=2.3cm]{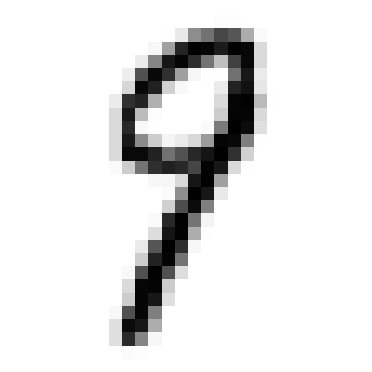} & \includegraphics[width=2.3cm]{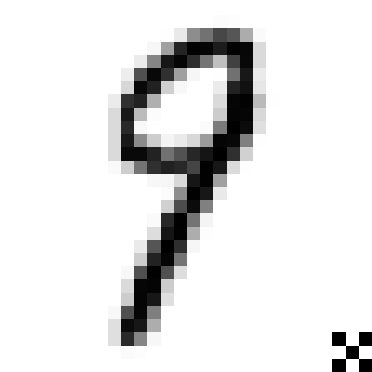} & \includegraphics[width=2.3cm]{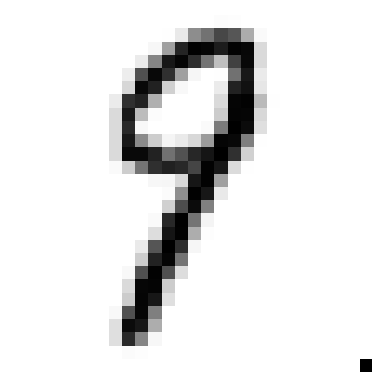} & \raisebox{0cm}[0pt][0pt]{\rotatebox[origin=c]{90}{EMNIST}} & \raisebox{-2.3cm}[0pt][0pt]{\rotatebox[origin=c]{90}{\textit{Dataset}}} \\
    
    \includegraphics[width=2.3cm]{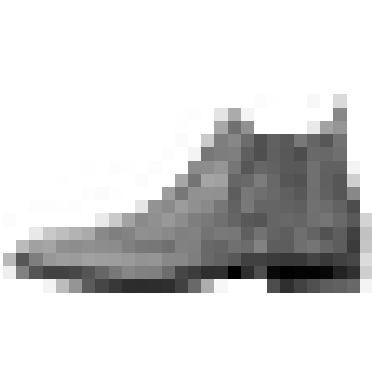} & \includegraphics[width=2.3cm]{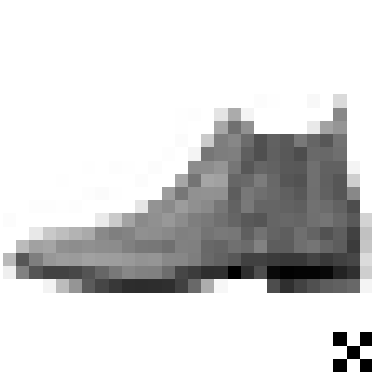} & \includegraphics[width=2.3cm]{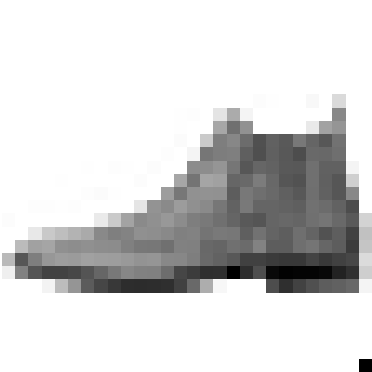} & \raisebox{0cm}[0pt][0pt]{\rotatebox[origin=c]{90}{Fashion MNIST}} &  \\
    \includegraphics[width=2.3cm]{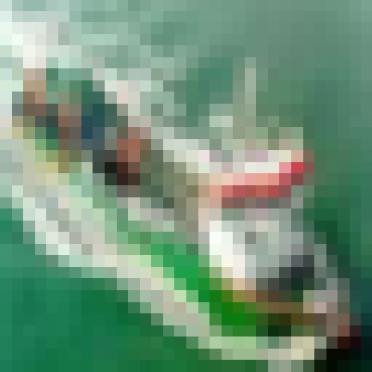} & \includegraphics[width=2.3cm]{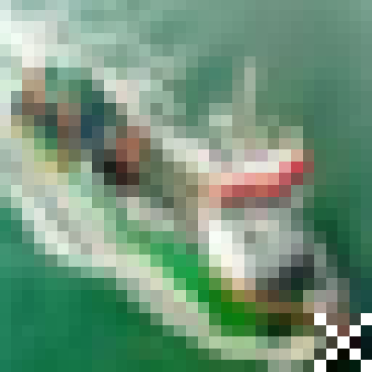} & \includegraphics[width=2.3cm]{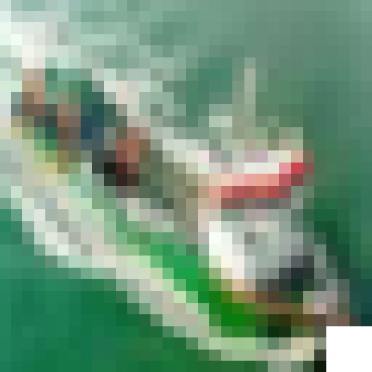} &\raisebox{0cm}[0pt][0pt]{\rotatebox[origin=c]{90}{CIFAR 10}} &  \\
    \centering  Original  & \centering  Cross & \centering Square & & \\
    \multicolumn{3}{c}{\textit{Pattern}} & & \\
    \end{tabular}}
    \caption{Examples of original and backdoored samples with cross patterns  and square patterns of EMNIST, Fashion-MNIST and CIFAR datasets.}
    \label{fig:patterns}
  \end{figure}

\nuria{Regarding the \ecml{backdoor} attack, we implement} pattern-key backdoor attacks, in which all the clients know the complete pattern and use it in their training process. We set a target label and a pattern key. The attack consists in classifying any sample poisoned with the pattern-key as the target label. \ecml{We employed a single static pattern for the attack, as this type of attack is generally more effective due to its collective nature targeting a common target~\cite{survey-nuria}.} In order to show that the behavior observed is agnostic of the pattern-key, we use two patterns \nuria{for each dataset} shown in Figure \ref{fig:patterns}: (1) a single black pixel for EMNIST and Fashion MNIST and a 5x5 white square for CIFAR-10, and (2) a black cross of length 3 for EMNIST and Fashion MNIST and a white cross of length 5 for CIFAR-10.

\vspace{-0.8em}
\subsection{Attack scenarios}\label{sec:scenarios}

\nuria{With the aim of testing the proposals in different configurations we consider two distinct scenarios:}
\begin{enumerate}
    \item \textbf{Scenario A:} There is only one adversarial client participating at a certain learning round $t$.
    \item \textbf{Scenario B: }The amount of attackers in a learning round is set to the number of miners in the network. In the case that the architecture being tested makes use of pooled-mining, that is, each miner get assigned a subset of the clients and make no use of any kind of resource from another miner's subset, we assure that each pool contains one adversarial client.
\end{enumerate}

\vspace{-0.8em}
\subsection{Evaluation metrics}\label{sec:metrics}
\ecml{Since the purpose of an untargeted attack is to degrade the performance of the model, we will consider the accuracy of the model in the corresponding test dataset of the model. On the other hand, when measuring a defense against a backdoor attack}, we must consider two aspects. The performance of the resulting model in the original task and in the \nuria{backdoor} task. The aim is to reduce the effects of the attack as much as possible without compromising the performance of the model in the original task. \nuria{For measuring this double goal we consider the following test datasets}:
\begin{enumerate}
    \item \emph{Original task test}. The original test of the dataset used for measuring the performance in terms of accuracy in the original task.
    \item \emph{Backdoor task test}. This dataset represents the attack in order to measure the performance of the backdoor task. Consists of the test instances but poisoned using the pattern in order to measure the capability of generalization of the attack.
\end{enumerate}

Since the results may be highly heterogeneous in each learning round and in order to show robust results, we offer two distinct metrics for each dataset: (1) the accuracy in the last learning round of the model in the dataset, which will be denoted by $accuracy$ and (2) the average of \textit{accuracy} throughout the last ten learning rounds, which will be denoted by $accuracy_{10}$. Also, Since results may differ from each miner in the case of PoFL and KFC, we show the metrics of the best miner in terms of performance in the original task, which we can measure with the \ecml{validation} dataset \ecml{used by the PoFL consensus mechanism}.

\vspace{-0.8em}
\subsection{Baselines}\label{sec:baselines}

We compare PoFL and KFC with the following FL architectures which represent the classical baselines and the most studied architectures for blockchain applied to FL, ecml{and the most common aggregation mechanisms} :
\begin{enumerate}
    \item \emph{Client-Server (C-S)}\cite{mcmahan-2017}. It makes no use of blockchain. It is one of the most common FL settings where one central server acts as an aggregator and schedules the whole learning process.
    \item \emph{Proof Of Work (PoW)}\cite{zhu-2023}. \ecml{A coupled architecture using PoW,} the most famous consensus mechanism where miners race each other in order to solve a computationally expensive puzzle. The winner will act as the aggregator for the given learning round.
    \item \emph{Proof Of Stake (PoS)}\cite{zhu-2023}. \ecml{A coupled architecture using PoS,} a general purpose consensus mechanism proposed as an energy efficient alternative to PoW where the selected miner is chosen on a deterministic fashion based on his stake or wealth on the network.
    \item \ecml{Krum\cite{blanchard-2017}. It makes no use of blockchain. It employs of the Krum aggregation operator presented previously.}
    \item \ecml{Trimmed-mean\cite{trimmedmean}. It makes no use of blockchain. It utilizes the trimmed mean operator, a statistical method that excludes a specified percentage of the most extreme data points from the calculation of the mean. By default, 10\% of the data points are removed from each end of the distribution.}
\end{enumerate}

In \ecml{C-S, PoW, PoS} and PoFL we used Federated Averaging (FedAvg)~\cite{mcmahan-2017} as a federated aggregation operator which is often considered as the default aggregation operator for FL and the most studied in literature. It is noteworthy that both PoS and PoW consensus mechanisms operate within the same underlying blockchain architecture. To facilitate a comprehensive understanding of the results, we present combined data from both PoS and PoW implementations and we refer to this baseline as PoW/S.

\vspace{-0.8em}
\subsection{Implementation details}\label{sec:details_imp}

\nuria{As the main purpose of the work is to defend against adversarial attacks, we make use of standard image classification deep learning models. We deploy an image classification deep learning model in the FL setting. We use an standard CNN-based image classification model composed of two CNN layers followed by its corresponding max-pooling layers, a dense layer and the output layer for the EMNIST and Fashion MNIST datasets and a pre-trained model based on EfficientNet \cite{efficientnet-2019} for the CIFAR-10 dataset.}
We provide the code used to run all the experiments in order to ensure reproducibility\footnote{\url{https://github.com/ari-dasci/S-kfc}}. \ecml{The code is written using the FLEXible FL framework \cite{herrera2024flex} and making use of his companion libraries \texttt{flex-block} and \texttt{flex-clash} for simulating the blockchain behavior and the attacks on the FL scheme respectively.}

\vspace{-0.8em}
\section{Analysis of PoFL as a defense mechanism}\label{sec:analysis_pofl}

In this section we analyze the performance of PoFL as a defense mechanism in Scenario A (see Section \ref{sec:pofl_a}) and Scenario B (see Section \ref{sec:pofl_b}). \ecml{We provide in this section the key information for understanding our findings. A more detailed visualization of our results can be found in Appendix \ref{sec:figures}}.

\vspace{-0.8em}
\subsection{Scenario A: just one miner being attacked}\label{sec:pofl_a}

In this section we analyze the performance of PoFL as a defense in Scenario A. The results shown in \ecml{Figure \ref{fig:pofl_a}} reveal a consistent trend throughout the experiments. We now delve into a detailed explanation of this observed pattern:

\begin{figure}[!ht]
\begin{subfigure}{0.32\linewidth}
  \centering
  \captionsetup{justification=centering}
  \includegraphics[width=\linewidth]{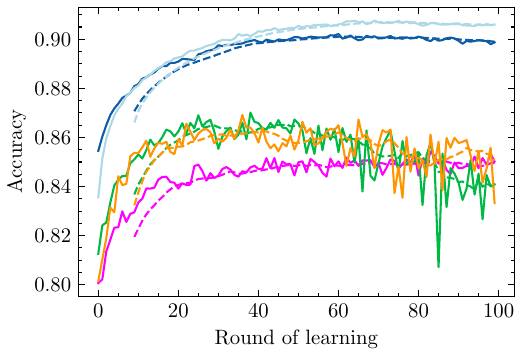}
  \caption{Fashion EMNIST\\ (original task).}
\end{subfigure} 
\begin{subfigure}{0.32\linewidth}
  \centering
  \captionsetup{justification=centering}
  \includegraphics[width=\linewidth]{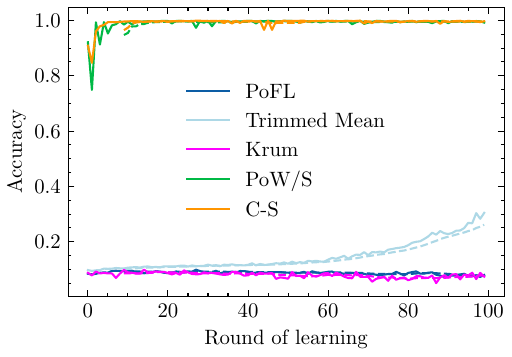}
  \caption{Fashion EMNIST\\ (backdoor task).}
\end{subfigure} 
\centering
\begin{subfigure}{0.32\linewidth}
  \centering
  \captionsetup{justification=centering}
  \includegraphics[width=\linewidth]{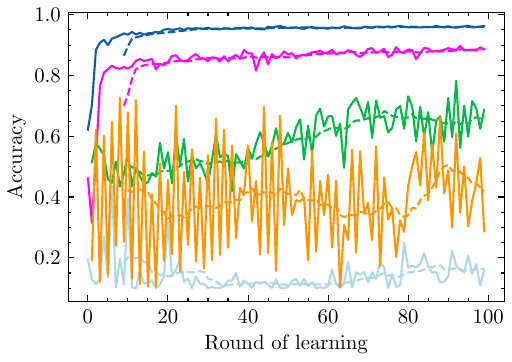}
  \caption{CIFAR\\ (Byzantine Attack).}
\end{subfigure}
\vskip\baselineskip
\caption{$accuracy$ (line) and $accuracy_{10}$ (dashed line) under the Backdoor attack on Fashion MNIST (a and b) and under the Byzantine attack on CIFAR (c) datasets, respectively, in \textbf{scenario A}.}
\label{fig:pofl_a}
\end{figure}

\begin{enumerate}
    \item \ecml{In the context of the Byzantine attack, PoFL consistently demonstrated a high performance. Conversely, the PoW/S and C-S architectures exhibited fluctuating accuracy levels and generally lower values compared to those achieved by PoFL, confirming the success of the Byzantine attack. Finally, Krum exhibited resistance but a lower performance than PoFL, which is an expected result of this operator. The Trimmed-mean approach, though effective in two of the datasets, exhibited a significant drop in performance on CIFAR, indicating its limitations as a defense strategy in certain scenarios.}
    \item For the backdoor attack, PoFL \ecml{and Trimmed-mean} demonstrates consistent superiority across original task test sets, achieving the highest accuracy throughout the training process. While the baseline approaches exhibit comparable accuracy levels, they consistently fall short of PoFL's performance.
    \item PoFL \ecml{and Krum} demonstrably mitigates the \ecml{backdoor attack}, achieving the a low accuracy on the backdoor task set. This minimal accuracy on the backdoor task suggests successful suppression of the malicious functionality. Conversely, client-server, blockchain architectures relying on PoW or PoS consensus mechanisms \ecml{and Trimmed-mean, in some cases,} exhibit no resistance. These baseline approaches achieve near-perfect accuracy on the secondary task, confirming the backdoor's successful integration into the federated models.
\end{enumerate}

Building upon these observations, the experimental results offer compelling evidence that PoFL constitutes a robust \nuria{FL} blockchain architecture against \ecml{adversarial} attacks within the investigated scenario, \ecml{showing similar or better results that other known methods such as Krum or Trimmed-mean}. These findings confirm that the performance-based consensus mechanism employed by PoFL, in conjunction with the pooled-mining architecture, effectively filters out malicious model updates injected by adversarial clients.

\vspace{-0.8em}
\subsection{Scenario B: all miners being attacked}\label{sec:pofl_b}

In this section we analyze the performance of PoFL as a defense in Scenario B. The results shown in the \ecml{Figure \ref{fig:pofl_b}} highlight the main weaknesses of PoFL. A closer examination of these results reveals the following observations:

\begin{figure}[!ht]
\begin{subfigure}{0.32\linewidth}
  \centering
  \captionsetup{justification=centering}
  \includegraphics[width=\linewidth]{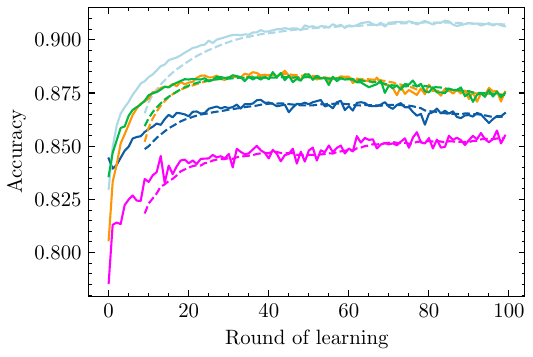}
  \caption{Fashion EMNIST\\ (original task).}
\end{subfigure} 
\begin{subfigure}{0.32\linewidth}
  \centering
  \captionsetup{justification=centering}
  \includegraphics[width=\linewidth]{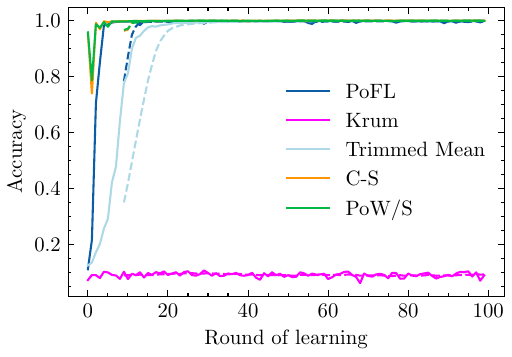}
  \caption{Fashion EMNIST\\ (backdoor task).}
\end{subfigure} 
\centering
\begin{subfigure}{0.32\linewidth}
  \centering
  \captionsetup{justification=centering}
  \includegraphics[width=\linewidth]{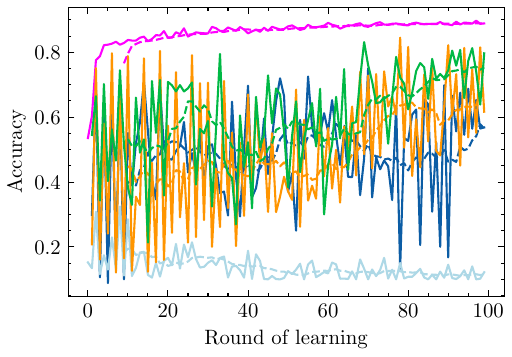}
  \caption{CIFAR \\(Byzantine Attack).}
\end{subfigure}
\vskip\baselineskip
\caption{$accuracy$ (line) and $accuracy_{10}$ (dashed line) under the Backdoor attack on Fashion MNIST (a and b) and under the Byzantine attack on CIFAR (c) datasets, respectively, in \textbf{scenario B}.}
\label{fig:pofl_b}
\end{figure}

\begin{enumerate}
    \item \ecml{In the context of the Byzantine attack, PoFL exhibited fluctuations during training similar to those observed in C-S or PoW/S during the previous scenario, accompanied by a decline in accuracy. This performance degradation indicates the susceptibility of the PoFL architecture to the Byzantine attack in this particular scenario. The Krum operator shows no decline indicating the success in supressing the attack.}
    
    \item For the backdoor attack, in contrast to the previous scenario where PoFL emerged as a high performing architecture, here it exhibits a weak performance in the original task. PoFL shows inconsistency with fluctuations throughout the training process. The considered baseline approaches, however, maintain a similar \ecml{behaviour compared to scenario A}, unlike PoFL's significant decline.

    \item Moreover, in this scenario, PoFL exhibits a concerning vulnerability. PoFL achieves high performance on the backdoor tasks, indicating success for the attack. While its accuracy is lower than the baselines, which remains consistent with the previous scenario, it remains concerningly high (over 80\% in average). Notably, even with fluctuations, PoFL exhibits a trend towards convergence on high accuracy for the backdoor task.
\end{enumerate}

In conclusion, the experiments paint a concerning picture for PoFL's suitability in this scenario. As evidenced by the previous observations, PoFL exhibits a significant vulnerability to \ecml{adversarial} attacks. Not only does it achieve concerningly high accuracy on the backdoor tasks, exceeding 80\% on average, but its performance on the original task \ecml{under both attacks} also suffers a substantial decline. 

This vulnerability of PoFL motivates the design of KFC, as the combination of PoFL and a defense mechanism in each miner making it resilient to adversarial attacks even when all miners are being attacked. We go through this question in the following section.

\vspace{-0.8em}
\section{Analysis of the performance of KFC \paco{Defense Strategy}}\label{sec:analysis_kfc}

In this section we analyze the performance of KFC as a defense against backdoor attacks in Scenario A (see Section \ref{sec:kfc_a}) and Scenario B (see Section \ref{sec:kfc_b}).

\subsection{Scenario A: just one miner being attacked}\label{sec:kfc_a}

\begin{figure}[!ht]
\begin{subfigure}{0.32\linewidth}
  \centering
  \captionsetup{justification=centering}
  \includegraphics[width=\linewidth]{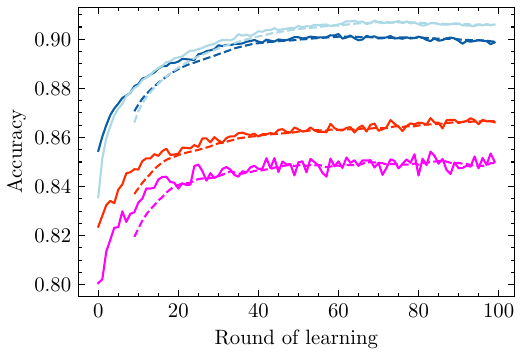}
  \caption{Fashion EMNIST \\(original task).}
\end{subfigure} 
\begin{subfigure}{0.32\linewidth}
  \centering
  \captionsetup{justification=centering}
  \includegraphics[width=\linewidth]{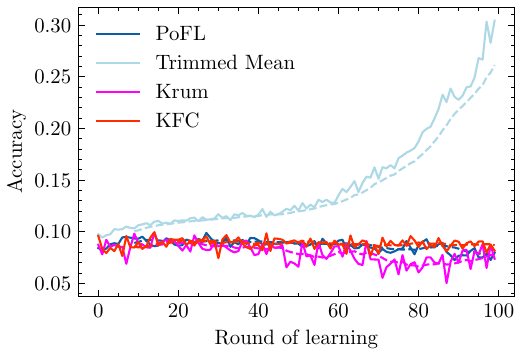}
  \caption{Fashion EMNIST \\(backdoor task).}
\end{subfigure} 
\centering
\begin{subfigure}{0.32\linewidth}
  \centering
  \captionsetup{justification=centering}
  \includegraphics[width=\linewidth]{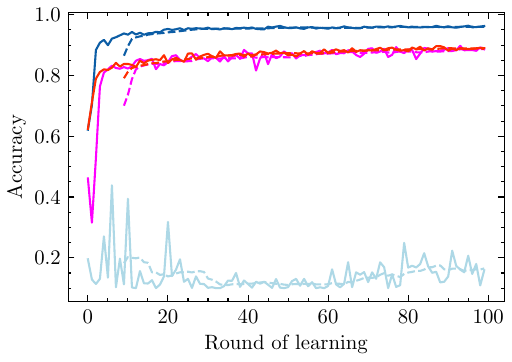}
  \caption{CIFAR \\(Byzantine Attack).}
\end{subfigure}
\vskip\baselineskip
\caption{$accuracy$ (line) and $accuracy_{10}$ (dashed line) under the Backdoor attack on Fashion MNIST (a and b) and under the Byzantine attack on CIFAR (c) datasets, respectively, in \textbf{scenario A}.}
\label{fig:kfc_a}
\end{figure}

If we analyze the results shown in \ecml{Figure \ref{fig:kfc_a}} we conclude that indeed the results obtained by KFC in scenario A, where PoFL was already an effective defense, are equivalent. If we take a more detailed inspection, there is a slight accuracy decrease \ecml{in the Byzantine attack and }in the original task due to this mechanism of selecting only the best client in each miner given by Krum. \ecml{Our analysis indicates that the KFC defense mechanism outperforms the Krum aggregation operator, upon which it is founded. These findings suggest that KFC is a viable and effective countermeasure against adversarial attacks in this particular context. The strategy consistently demonstrated satisfactory performance in mitigating the impact of such attacks.}

\vspace{-0.8em}
\subsection{Scenario B: all miners being attacked}\label{sec:kfc_b}

Considering, again, the more challenging scenario where the existence of a pool devoid of malicious clients is no longer a guaranteed assumption, we now focus on \ecml{Figure \ref{fig:kfc_b}}. A more detailed analysis of these findings highlights the following points:

\begin{figure}[!ht]
\begin{subfigure}{0.32\linewidth}
  \centering
  \captionsetup{justification=centering}
  \includegraphics[width=\linewidth]{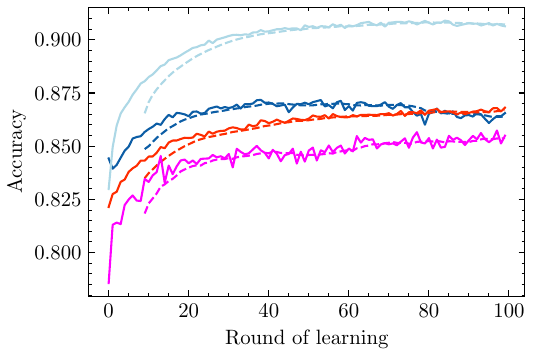}
  \caption{Fashion EMNIST \\(original task).}
\end{subfigure} 
\begin{subfigure}{0.32\linewidth}
  \centering
  \captionsetup{justification=centering}
  \includegraphics[width=\linewidth]{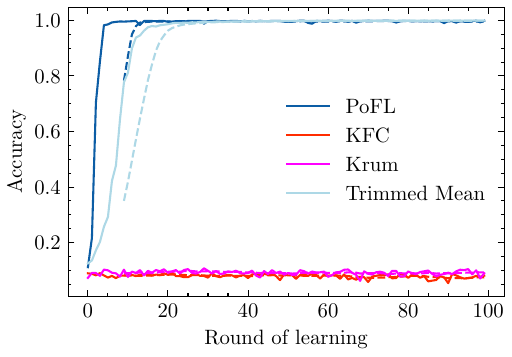}
  \caption{Fashion EMNIST \\(backdoor task).}
\end{subfigure} 
\centering
\begin{subfigure}{0.32\linewidth}
  \centering
  \captionsetup{justification=centering}
  \includegraphics[width=\linewidth]{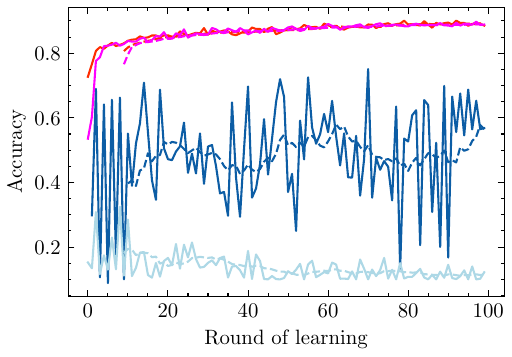}
  \caption{CIFAR \\(Byzantine Attack).}
\end{subfigure}
\vskip\baselineskip
\caption{$accuracy$ (line) and $accuracy_{10}$ (dashed line) under the Backdoor attack on Fashion MNIST (a and b) and under the Byzantine attack on CIFAR (c) datasets, respectively, in \textbf{scenario B}.}
\label{fig:kfc_b}
\end{figure}

\begin{enumerate}
    \item \ecml{In the context of the Byzantine attack, KFC exhibited a performance similar to the previous scenario, maintaining a consistently high accuracy across all datasets, in contrast to PoFL. Consequently, it can be concluded that KFC effectively mitigated the Byzantine attack under these conditions.}
    
    \item \ecml{For the backdoor attack}, KFC's ability to maintain accuracy in the original task compared to PoFL demonstrates its resilience in the new challenging scenario where all miners are under attack.

    \item Conversely to PoFL, KFC demonstrates exceptional resilience, achieving minimal accuracy on the backdoor task, effectively mitigating the attack.
\end{enumerate}

To sum up, KFC stands out as defense mechanism. It demonstrates exceptional resilience against \ecml{adversarial} attacks, achieving minimal accuracy on the backdoor task and maintaining strong performance on both the original task \ecml{and under the Byzantine attack}, outperforming PoFL, \ecml{Trimmed-mean and the Krum operator}. These findings strongly confirm that KFC surpasses PoFL as the preferred choice in scenarios where the pool might be compromised. KFC effectively achieves \ecml{both mitigating the Byzantine attack and} the two-fold objective of defending against backdoor attacks in \nuria{FL}: maximizing the original task's performance and minimizing the impact of potential backdoor attacks.

\section{\paco{Discussion}}\label{sec:limitations}
\ecml{Our experimental results offer robust evidence in support of KFC as a defensive mechanism. Nevertheless, to uphold the scientific rigor of our investigation and ensure the integrity of our conclusions, it is essential to acknowledge potential methodological limitations. By openly discussing these limitations, we seek to provide a more refined interpretation of our findings:}

\begin{itemize}
    \item \ecml{\textbf{Computational resources.} The Krum operator employed in this study exhibits a quadratic time complexity ($O(n^2 \cdot (d + \log{n}))$), which can be computationally inefficient, particularly for large sets of clients. However, the proposed KFC algorithm leverages pooled mining, which in some cases might mitigate this limitation. By proportionally increasing the number of miners in relation to the client count during each training round, the computational burden can be effectively distributed, ensuring scalability even for substantial numbers of participants. Furthermore, investigating the feasibility of substituting the Krum operator in KFC with a more efficient alternative constitutes an important area for future research.}

    \item \ecml{\textbf{Communication Overhead.} The present investigation did not delve into network-related aspects such as latency or communication overhead during the experimental evaluation. However, addressing these issues is a growing area of research, as evidenced by recent studies~\cite{qu2022}. Considering KFC's reliance on the Krum operator and a widely adopted consensus mechanism like PoFL, the integration of techniques designed to mitigate communication overhead within the KFC framework appears promising.}

    \item \ecml{\textbf{Complexity of Backdoor attacks.} The present study implemented a pattern-key attack with a single pattern and static accessory injection, a relatively straightforward backdoor attack technique~\cite{survey-nuria}. However, more sophisticated and stealthy backdoor attacks have been devised to circumvent multiple defense mechanisms in federated learning. While the implemented attack is considered one of the most successful due to its targeted approach, the effectiveness of KFC as a defense against more advanced backdoor attacks might require further investigation.}
\end{itemize}

\section{Conclusions and future work} \label{sec:conclusion}

In this paper we study the potential of integrating blockchain technology with FL to defend against adversarial attacks, in particular \ecml{Byzantine and} backdoor attacks. We test the hypothesis that PoFL, a consensus mechanism designed ad-hoc for FL, which could be resilient to adversarial attacks in some situations. So, we propose KFC, a defense mechanism based on the combination of Krum and PoFL which aims to defend the federated scheme under any configuration of attacks. The conclusions obtained from the experimental results are:

\begin{itemize}
    \item PoFL defend against \ecml{Byzantine and} backdoor attacks in FL when just one miner is being attacked. In fact, it outperforms other consensus mechanisms considered as baselines such as PoW, PoS, \ecml{Krum and Trimmed-mean in both maintaining original task performance under Byzantine and backdoor attacks and mitigating the effects of these attacks}.
    
    \item PoFL is not enough to defend against \ecml{adversarial} attacks when all miners are being attacked.
    
    \item KFC is able to defend against \ecml{adversarial} attacks even when all the miners are being attacked. It success in maintaining \ecml{the performance under the Byzantine attack and} the two-fold objective regarding backdoor attacks: the best performance in the original task, and the worst performance of the backdoor task (the attacked is not being injected).
\end{itemize}

In conclusion, the blockchain has shown to be a feasible solution to the vulnerability of FL to adversarial attacks. However, to be resilient to all scenarios of attack, it may need to be combined with other defense techniques. As future work, we plan to apply this combination of blockchain and PoFL for FL resilience against more types of attacks, including \ecml{more complex backdoor attacks} and privacy attacks.

\section{Acknowledgments}
This research results from the Strategic Project IAFER-Cib (C074/23), as a result of the collaboration agreement signed between the National Institute of Cybersecurity (INCIBE) and the University of Granada. This initiative is carried out within the framework of the Recovery, Transformation and Resilience Plan funds, financed by the European Union (Next Generation).

\bibliographystyle{unsrt}  
\bibliography{references}  

\appendix

\section{Experiments Results Visualization}\label{sec:figures}

The supplementary figures and tables included in this appendix serve to corroborate the experimental data presented in the main text. These figures provide a detailed and comprehensive representation of the results, facilitating a more in-depth analysis of the findings.

\begin{table*}[!ht]
\resizebox{\linewidth}{!}{
\centering
\begin{tabular}{llrrrlllrrrll}
\toprule
 & & \multicolumn{5}{c}{\textit{accuracy}} &&\multicolumn{5}{c}{\textit{$accuracy_{10}$}}\\
\toprule
       &          & \textbf{C-S} & \textbf{PoW/S} & \textbf{PoFL}  & \textbf{Trim.-}&\textbf{Krum} && \textbf{C-S} & \textbf{PoW/S} & \textbf{PoFL}   & \textbf{Trim.-}& \textbf{Krum}\\
 & & & & & \textbf{mean}&  && & & & \textbf{mean}&\\ 
\toprule
\textbf{EMNIST} & Original & 0.982& 0.979& \textbf{0.993}& 0.993&0.986 && 0.982& 0.980& \textbf{0.993}& 0.993&0.988\\  
       & Backdoor & 0.999& 0.999& 0.098& 0.999&\textbf{0.097} && 0.999& 0.999& \textbf{0.097}& 0.999&0.097\\  
\midrule
\textbf{Fashion}   & Original & 0.846& 0.841& 0.896& \textbf{0.906}&0.854 && 0.841& 0.848& 0.898& \textbf{0.907}&0.853\\ 
 \textbf{MNIST}    & Backdoor & 0.995& 0.996& \textbf{0.087}& 0.999&0.089 && 0.988& 0.987& 0.098& 0.999&\textbf{0.092}\\ 
\midrule
\textbf{CIFAR-10}        & Original & 0.910& 0.903& \textbf{0.939}& 0.887&0.882 && 0.902& 0.876& \textbf{0.928}& 0.886&0.883\\ 
                & Backdoor & 0.967& 0.970& 0.116& 0.087&\textbf{0.086} && 0.934& 0.973& \textbf{0.084}& 0.088&0.085\\ 
\bottomrule
\end{tabular}}
\caption{\textit{accuracy} and \textit{$accuracy_{10}$} comparing PoFL with the baselines in the \textbf{scenario A} under a backdoor attack.}\label{tab:poflbackdoor_a}
\end{table*}

\begin{table*}[!ht]
\resizebox{\linewidth}{!}{
\centering
\begin{tabular}{lrrrlllrrrll}
\toprule
 & \multicolumn{5}{c}{\textit{accuracy}} && \multicolumn{5}{c}{\textit{$accuracy_{10}$}}\\
\toprule
       & \textbf{C-S} & \textbf{PoW/S} & \textbf{PoFL}  & \textbf{Trim.-}&\textbf{Krum} && \textbf{C-S} & \textbf{PoW/S} & \textbf{PoFL}   & \textbf{Trim.-}&\textbf{Krum}\\
 & & & & \textbf{mean}&  && & & & \textbf{mean}&\\ 
\toprule
\textbf{EMNIST} & 0.893& 0.100& 0.992& \textbf{0.993}&0.985 && 0.659& 0.103& 0.992& \textbf{0.993}&0.985\\  
\midrule
\textbf{Fashion MNIST}& 0.826& 0.104& 0.905& \textbf{0.905}&0.853 && 0.836& 0.097& 0.904& \textbf{0.906}&0.853\\ 
\midrule
\textbf{CIFAR-10}        & 0.286& 0.687& \textbf{0.962}& 0.158&0.886 && 0.419& 0.668& \textbf{0.960}& 0.165&0.8866\\
\bottomrule
\end{tabular}}
\caption{\textit{accuracy} and \textit{$accuracy_{10}$} comparing PoFL with the baselines in the \textbf{scenario A} under a Byzantine attack.}\label{tab:poflbyzantine_a}

\end{table*}

\begin{figure}[!ht]
\begin{subfigure}{0.49\linewidth}
  \centering
  \includegraphics[width=\linewidth]{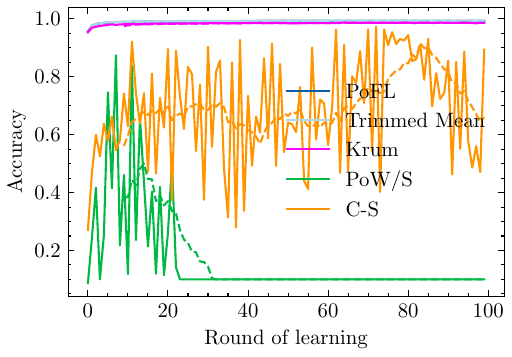}
  \caption{EMNIST.}
\end{subfigure} 
\begin{subfigure}{0.49\linewidth}
  \centering
  \includegraphics[width=\linewidth]{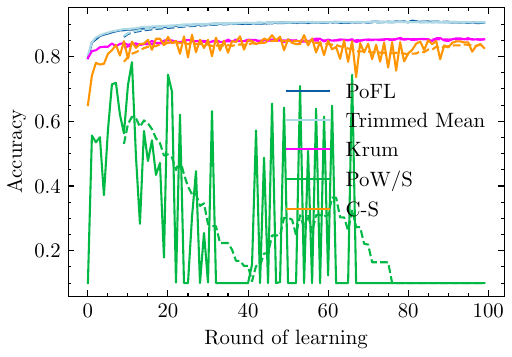}
  \caption{Fashion EMNIST.}
\end{subfigure} 
\vskip\baselineskip
\centering
\begin{subfigure}{0.49\linewidth}
  \centering
  \includegraphics[width=\linewidth]{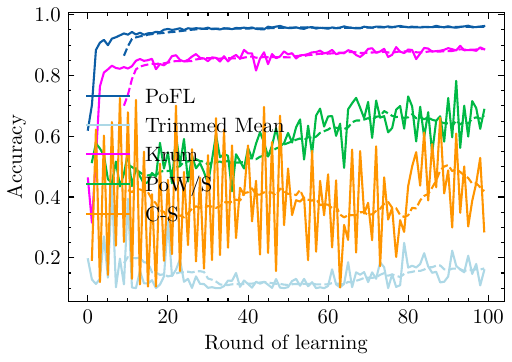}
  \caption{CIFAR.}
\end{subfigure}
\vskip\baselineskip
\caption{$accuracy$ (line) and $accuracy_{10}$ (dashed line) under the Byzantine attack on EMNIST (a), Fashion MNIST (b) and CIFAR (c) datasets, respectively, in \textbf{scenario A}.}
\end{figure}

\begin{figure}[!ht]
\begin{subfigure}{0.49\linewidth}
  \centering
  \includegraphics[width=\linewidth]{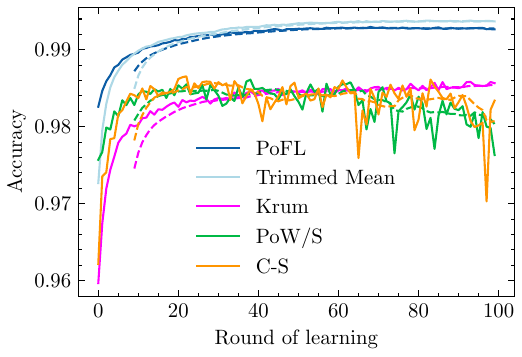}
  \caption{EMNIST (Original task).}
\end{subfigure} 
\begin{subfigure}{0.49\linewidth}
  \centering
  \includegraphics[width=\linewidth]{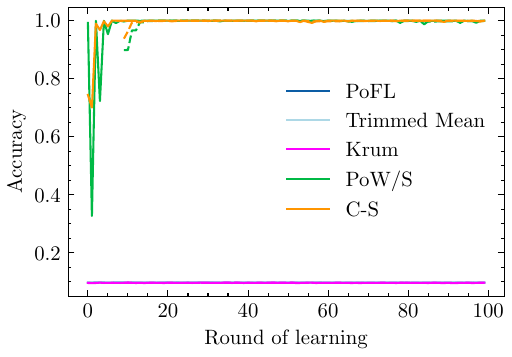}
  \caption{EMNIST (Backdoor task).}
\end{subfigure} 
\vskip\baselineskip
\begin{subfigure}{0.49\linewidth}
  \centering
  \includegraphics[width=\linewidth]{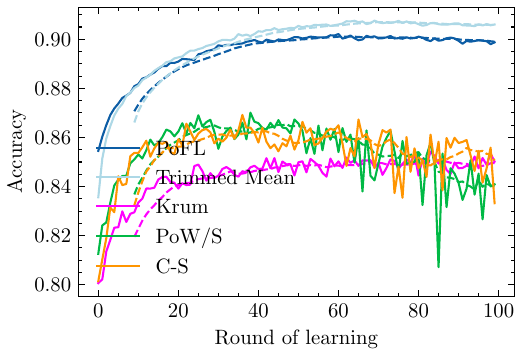}
  \caption{Fashion (Original task).}
\end{subfigure} 
\begin{subfigure}{0.49\linewidth}
  \centering
  \includegraphics[width=\linewidth]{backdoor_fashion_one.pdf}
  \caption{Fashion (Backdoor task).}
\end{subfigure}
\vskip\baselineskip
\begin{subfigure}{0.49\linewidth}
  \centering
  \includegraphics[width=\linewidth]{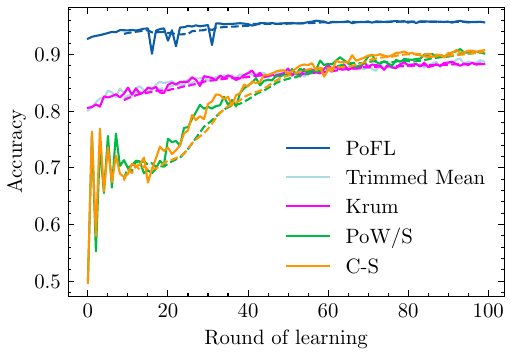}
  \caption{CIFAR (Original task).}
\end{subfigure}
\begin{subfigure}{0.49\linewidth}
  \centering
  \includegraphics[width=\linewidth]{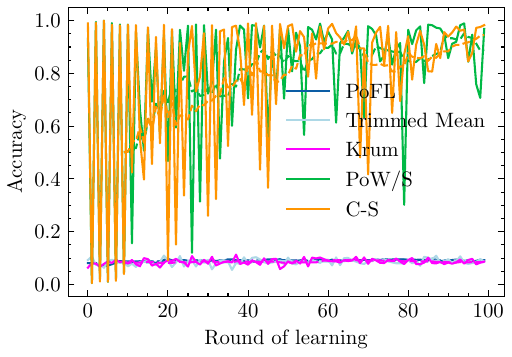}
  \caption{CIFAR (Backdoor task).}
\end{subfigure}
\vskip\baselineskip
\caption{$accuracy$ (line) and $accuracy_{10}$ (dashed line) of the original and backdoor tasks on EMNIST (a and b), Fashion MNIST (c and d) and CIFAR (e and f) datasets, respectively, in \textbf{scenario A}.}
\end{figure}

\begin{table}[h!]
    \resizebox{\linewidth}{!}{
    \centering
    \begin{tabular}{llccclllcccll}
    	\toprule
           &&\multicolumn{5}{c}{\textit{accuracy}} &&  \multicolumn{5}{c}{\textit{$accuracy_{10}$}}\\
           \toprule
           &&\textbf{C-S} &  \textbf{PoW/S} &  \textbf{PoFL}  & \textbf{Trim.-}&\textbf{Krum} &&  \textbf{C-S} &  \textbf{PoW/S} & \textbf{PoFL} 
 & \textbf{Trim.-}&\textbf{Krum}\\
 & & & & & \textbf{mean}&  && & & & \textbf{mean}&\\
\toprule
           \textbf{EMNIST} &Original 
&0.986&  0.988&  0.898& \textbf{0.993}&0.985 &&  0.986&  0.988& 0.910& \textbf{0.993}&0.985\\
           &Backdoor 
&0.999&  1.000&  1.000& 0.099&\textbf{0.096} &&  0.999&  1.000& 0.999& 0.099&\textbf{0.096}\\
\midrule
           \textbf{Fashion}   &Original 
&0.875&  0.874&  0.813& \textbf{0.906}&0.850 &&  0.874&  0.874& 0.800& \textbf{0.905}&0.849\\
           \textbf{MNIST}    &Backdoor 
&0.999&  0.999&  0.819& 0.304&\textbf{0.073} &&  0.999&  0.999& 0.918& 0.261&\textbf{0.077}\\
\midrule
           \textbf{CIFAR-10}        &Original 
&0.910&  \textbf{0.943}&  0.848& 0.886&0.883 &&  0.903&  \textbf{0.940}& 0.822& 0.885&0.883\\
           &Backdoor &0.967&  0.992&  0.853& 0.082&\textbf{0.067} &&  0.983&  0.992& 0.915& 0.091&\textbf{0.087}\\
           \bottomrule
    \end{tabular}}
    \caption{\textit{accuracy} and \textit{$accuracy_{10}$} comparing PoFL with the baselines in the \textbf{scenario B} under a backdoor attack.}\label{tab:poflbackdoor_b}
\end{table}

\begin{table}[h!]
\resizebox{\linewidth}{!}{
    \centering
    \begin{tabular}{lccclllcccll}
    	\toprule
           &\multicolumn{5}{c}{\textit{accuracy}} &&  \multicolumn{5}{c}{\textit{$accuracy_{10}$}}\\
           \toprule
           &\textbf{C-S} &  \textbf{PoW/S} &  \textbf{PoFL}  & \textbf{Trim.-}&\textbf{Krum} &&  \textbf{C-S} &  \textbf{PoW/S} & \textbf{PoFL} 
 & \textbf{Trim.-}&\textbf{Krum}\\
 & & & & \textbf{mean}&  && & & & \textbf{mean}&\\
\toprule
           \textbf{EMNIST} &0.923&  0.735&  0.920& \textbf{0.989}&0.985 &&  0.777&  0.724& 0.912& \textbf{0.989}&0.986\\
           \midrule
           \textbf{Fashion MNIST}&0.832&  0.737&  0.734& \textbf{0.884}&0.850 &&  0.822&  0.808& 0.747& \textbf{0.884}&0.851\\
           \midrule
           \textbf{CIFAR-10}        &0.616&  0.796&  0.564& 0.122&\textbf{0.888} &&  0.669&  0.754& 0.564& 0.115&\textbf{0.889}\\
           \bottomrule
    \end{tabular}}
    \caption{\textit{accuracy} and \textit{$accuracy_{10}$} comparing PoFL with the baselines in the \textbf{scenario B} under a Byzantine attack.}\label{tab:poflbyzantine_b}
\end{table}
\begin{figure}[h!]
\begin{subfigure}{0.49\linewidth}
  \centering
  \includegraphics[width=\linewidth]{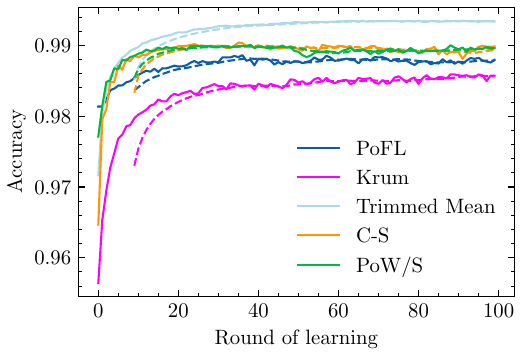}
  \caption{EMNIST (Original task).}
\end{subfigure} 
\begin{subfigure}{0.49\linewidth}
  \centering
  \includegraphics[width=\linewidth]{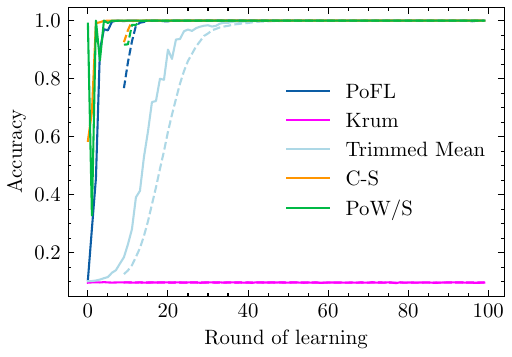}
  \caption{EMNIST (Backdoor task).}
\end{subfigure} 
\vskip\baselineskip
\begin{subfigure}{0.49\linewidth}
  \centering
  \includegraphics[width=\linewidth]{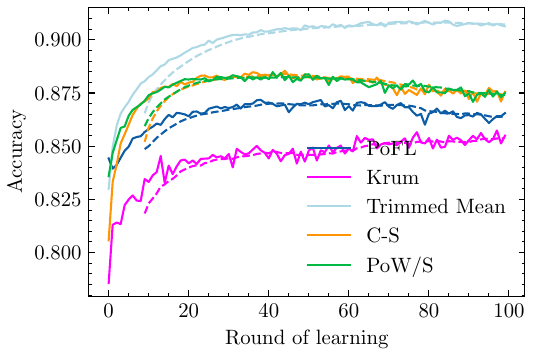}
  \caption{Fashion (Original task).}
\end{subfigure} 
\begin{subfigure}{0.49\linewidth}
  \centering
  \includegraphics[width=\linewidth]{backdoor_fashion_all.pdf}
  \caption{Fashion (Backdoor task).}
\end{subfigure}
\vskip\baselineskip
\begin{subfigure}{0.49\linewidth}
  \centering
  \includegraphics[width=\linewidth]{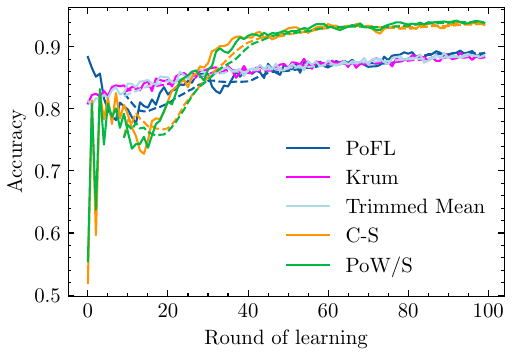}
  \caption{CIFAR (Original task).}
\end{subfigure}
\begin{subfigure}{0.49\linewidth}
  \centering
  \includegraphics[width=\linewidth]{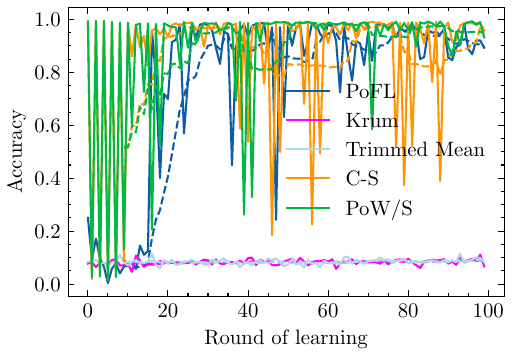}
  \caption{CIFAR (Backdoor task).}
\end{subfigure}
\vskip\baselineskip
\caption{$accuracy$ (line) and $accuracy_{10}$ (dashed line) of the original and backdoor tasks on EMNIST (a and b), Fashion MNIST (c and d) and CIFAR (e and f) datasets, respectively, in \textbf{scenario B}.}
\end{figure}

\begin{figure}[!ht]
\begin{subfigure}{0.49\linewidth}
  \centering
  \includegraphics[width=\linewidth]{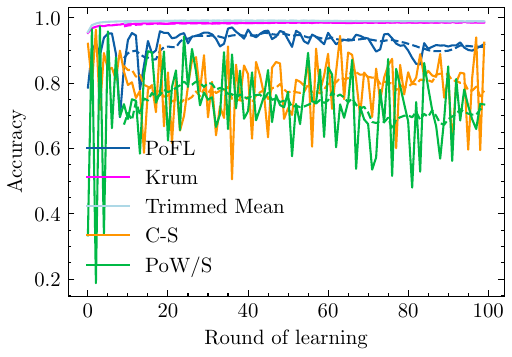}
  \caption{EMNIST.}
\end{subfigure} 
\begin{subfigure}{0.49\linewidth}
  \centering
  \includegraphics[width=\linewidth]{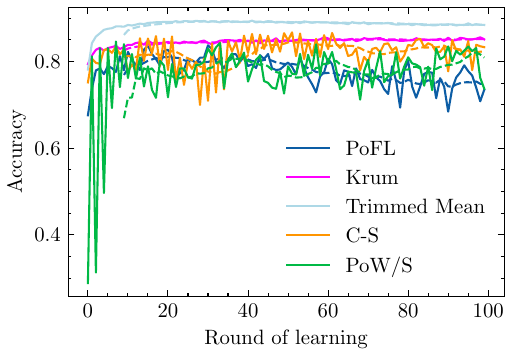}
  \caption{Fashion EMNIST.}
\end{subfigure} 
\vskip\baselineskip
\centering
\begin{subfigure}{0.49\linewidth}
  \centering
  \includegraphics[width=\linewidth]{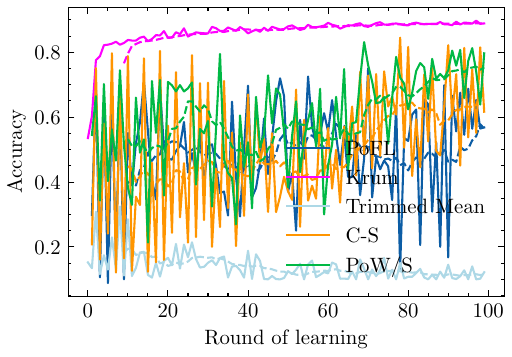}
  \caption{CIFAR.}
\end{subfigure}
\vskip\baselineskip
\caption{$accuracy$ (line) and $accuracy_{10}$ (dashed line) under the Byzantine attack on EMNIST (a), Fashion MNIST (b) and CIFAR (c) datasets, respectively, in \textbf{scenario B}.}
\end{figure}

\begin{table*}[h!]
\resizebox{\linewidth}{!}{
\centering
\begin{tabular}{llrrlllrrll}
\toprule
 & & \multicolumn{4}{c}{\textit{accuracy}} & & \multicolumn{4}{c}{\textit{$accuracy_{10}$}}\\
 \toprule
\textbf{}     &          & \textbf{PoFL}    & \textbf{KFC}  & \textbf{Trim.-}&\textbf{Krum} && \textbf{PoFL}    & \textbf{KFC}  & \textbf{Trim.-}&\textbf{Krum}\\
 & & & & \textbf{mean}&  && & & \textbf{mean}&\\ 
\midrule
\textbf{EMNIST}        & Original &   \textbf{0.993}&  0.988& 0.993&0.986 && \textbf{0.993}& 0.988& 0.993&0.988\\ 
              & Backdoor &     0.098& \textbf{0.096}& 0.999&0.097 &&    0.097&     \textbf{ 0.097}& 0.999&0.097\\ 
\midrule
\textbf{Fashion} & Original &    0.896& 0.864& \textbf{0.906}&0.854 &&  0.898&   0.864& \textbf{0.907}&0.853\\ 
\textbf{MNIST}  & Backdoor &   \textbf{0.079}& 0.081& 0.999&0.089 &&  \textbf{0.077}&  0.086& 0.999&0.092\\ 
\midrule
\textbf{CIFAR-10}      & Original & \textbf{0.956}& 0.887& 0.887&0.882 && \textbf{0.957}& 0.883& 0.886&0.883\\ 
                        & Backdoor &  0.092& 0.096& 0.087&\textbf{0.086} && 0.090&  \textbf{0.079}& 0.088&0.085\\ 
\bottomrule
\end{tabular}}
    \caption{\textit{accuracy} and $accuracy_{10}$ comparing \ecml{the baselines} and KFC in the \textbf{scenario A} under a backdoor attack.} \label{tab:kfcbackdoor_a}
\end{table*}

\begin{table*}[h!]
\resizebox{\linewidth}{!}{
\centering
\begin{tabular}{lrrlllrrll}
\toprule
 & \multicolumn{4}{c}{\textit{accuracy}} & & \multicolumn{4}{c}{\textit{$accuracy_{10}$}}\\
 \toprule
\textbf{}     & \textbf{PoFL}    & \textbf{KFC}  &  \textbf{Trim.-}&\textbf{Krum} && \textbf{PoFL}    & \textbf{KFC}  & \textbf{Trim.-}&\textbf{Krum}\\
 & & & \textbf{mean}&  && & & \textbf{mean}&\\ 
\midrule
\textbf{EMNIST}        &   0.992&  0.988&  \textbf{0.993}&0.985 && 0.992& 0.988& \textbf{0.993}&0.985\\ 
\midrule
\textbf{Fashion MNIST}&    0.905& 0.870&  \textbf{0.905}&0.853 &&  0.904&   0.868& \textbf{0.906}&0.853\\ 
\midrule
\textbf{CIFAR-10}      & \textbf{0.962}& 0.890&  0.158&0.886 && \textbf{0.960}& 0.887& 0.165&0.886\\
\bottomrule
\end{tabular}}
    \caption{\textit{accuracy} and $accuracy_{10}$ comparing \ecml{the baselines} and KFC in the \textbf{scenario A} under a Byzantine attack.}  \label{tab:kfcbyzantine_a}
\end{table*}

\begin{figure}[!ht]
\begin{subfigure}{0.49\linewidth}
  \centering
  \includegraphics[width=\linewidth]{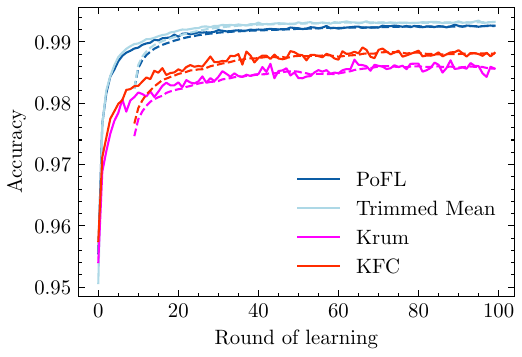}
  \caption{EMNIST.}
\end{subfigure} 
\begin{subfigure}{0.49\linewidth}
  \centering
  \includegraphics[width=\linewidth]{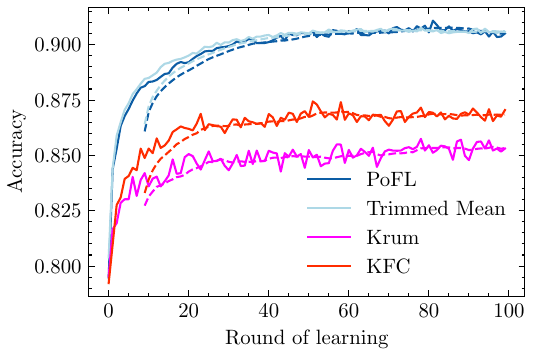}
  \caption{Fashion EMNIST.}
\end{subfigure} 
\vskip\baselineskip
\centering
\begin{subfigure}{0.49\linewidth}
  \centering
  \includegraphics[width=\linewidth]{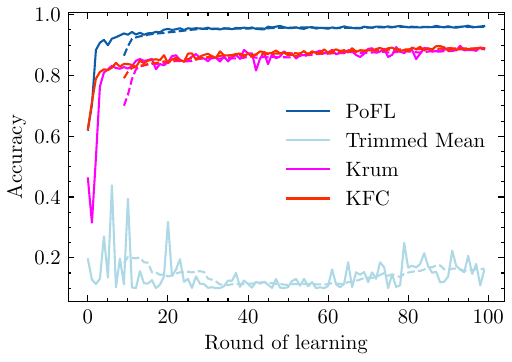}
  \caption{CIFAR.}
\end{subfigure}
\vskip\baselineskip
\caption{$accuracy$ (line) and $accuracy_{10}$ (dashed line) under the Byzantine attack on EMNIST (a), Fashion MNIST (b) and CIFAR (c) datasets, respectively, in \textbf{scenario A}.}
\end{figure}

\begin{figure}[h!]
\begin{subfigure}{0.49\linewidth}
  \centering
  \includegraphics[width=\linewidth]{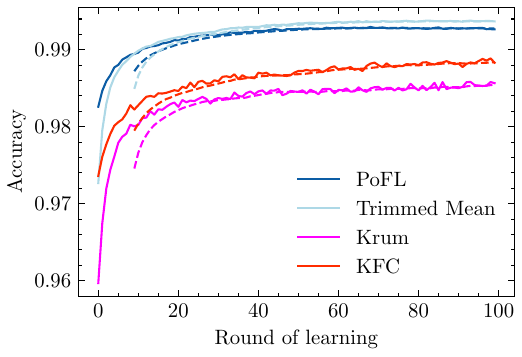}
  \caption{EMNIST (Original task).}
\end{subfigure} 
\begin{subfigure}{0.49\linewidth}
  \centering
  \includegraphics[width=\linewidth]{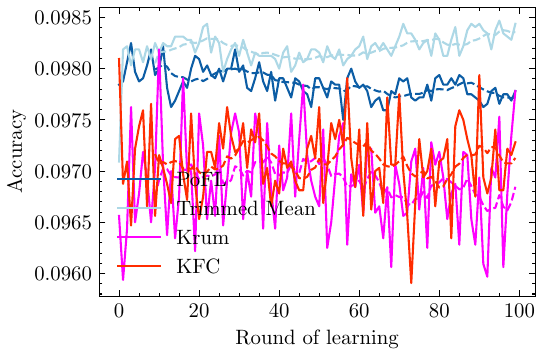}
  \caption{EMNIST (Backdoor task).}
\end{subfigure} 
\vskip\baselineskip
\begin{subfigure}{0.49\linewidth}
  \centering
  \includegraphics[width=\linewidth]{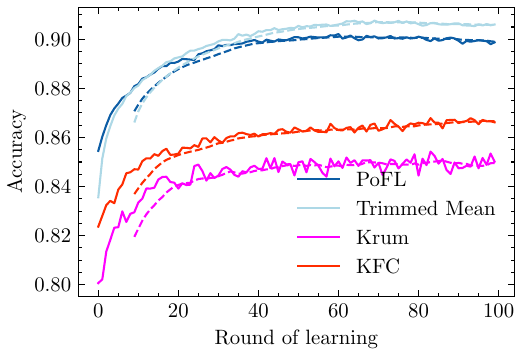}
  \caption{Fashion (Original task).}
\end{subfigure} 
\begin{subfigure}{0.49\linewidth}
  \centering
  \includegraphics[width=\linewidth]{kfc_vs_pofl_backdoor_fashion_one.pdf}
  \caption{Fashion (Backdoor task).}
\end{subfigure}
\vskip\baselineskip
\begin{subfigure}{0.49\linewidth}
  \centering
  \includegraphics[width=\linewidth]{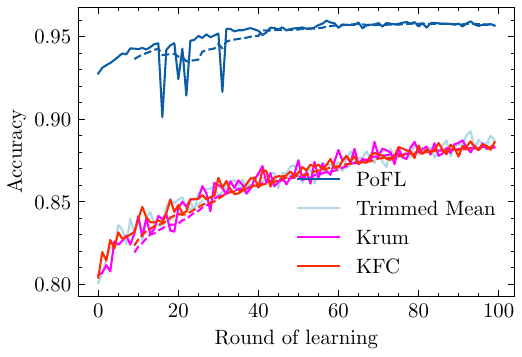}
  \caption{CIFAR (Original task).}
\end{subfigure}
\begin{subfigure}{0.49\linewidth}
  \centering
  \includegraphics[width=\linewidth]{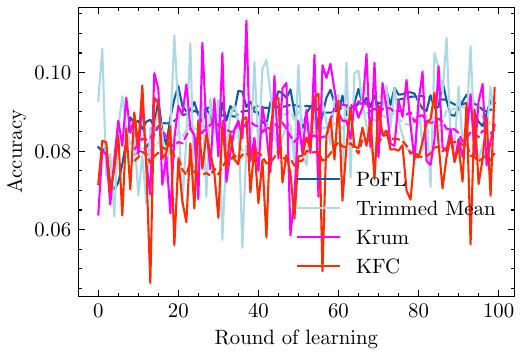}
  \caption{CIFAR (Backdoor task).}
\end{subfigure}
\vskip\baselineskip
\caption{$accuracy$ (line) and $accuracy_{10}$ (dashed line) of the original and backdoor tasks on EMNIST (a and b), Fashion MNIST (c and d) and CIFAR (e and f) datasets, respectively, in \textbf{scenario A}.}
\end{figure}

\begin{table}
\resizebox{\linewidth}{!}{
    \centering
    \begin{tabular}{llcclllccll}
    	\toprule
            &&\multicolumn{4}{c}{\textit{accuracy}} & &  \multicolumn{4}{c}{\textit{$accuracy_{10}$}
}\\
\toprule
            &&\textbf{PoFL}    &  \textbf{KFC}  &  \textbf{Trim.-}&\textbf{Krum} &&  \textbf{PoFL}& \textbf{KFC}
 &  \textbf{Trim.-}&\textbf{Krum}\\
 & & & & \textbf{mean}&  && & & \textbf{mean}&\\
\toprule
            \textbf{EMNIST}        &Original 
&0.898&  0.992&  \textbf{0.993}&0.985 &&  0.910& \textbf{0.992}&  0.991&0.985\\
            &Backdoor 
&1.000&  \textbf{0.096}&  0.099&0.096 &&  0.999& \textbf{0.096}&  0.099&0.096\\
\midrule
            \textbf{Fashion} &Original 
&0.863&  0.868&  \textbf{0.906}&0.850 &&  0.863& 0.866&  \textbf{0.905}&0.849\\
            \textbf{MNIST}  &Backdoor 
&0.998&  0.081&  0.304&\textbf{0.073} &&  0.996& \textbf{0.075}&  0.261&0.077\\
\midrule
            \textbf{CIFAR-10}      &Original 
&\textbf{0.889}&  0.885&  0.886&0.883 &&  0.822& 0.884&  0.885&0.883\\
            &Backdoor &0.897&  0.096&  0.082&\textbf{0.067} &&  0.897& \textbf{0.079}&  0.091&0.087\\
\bottomrule
    \end{tabular}}
     \caption{\textit{accuracy} and $accuracy_{10}$ comparing \ecml{the baselines} and KFC in the \textbf{scenario B} under a backdoor attack.}
    \label{tab:kfcbackdoorb}
\end{table}

\begin{table*}[h!]
\resizebox{\linewidth}{!}{
\centering
\begin{tabular}{lrrlllrrll}
\toprule
 & \multicolumn{4}{c}{\textit{accuracy}} &  & \multicolumn{4}{c}{\textit{$accuracy_{10}$}}  \\
 \toprule
\textbf{}     & \textbf{PoFL}    & \textbf{KFC}  &  \textbf{Trim.-} & \textbf{Krum}  && \textbf{PoFL}    & \textbf{KFC}  &  \textbf{Trim.-} & \textbf{Krum} \\
 & & & \textbf{mean} &  && & & \textbf{mean} & \\ 
\midrule
\textbf{EMNIST}        &   0.920&  0.988&  \textbf{0.989} & 0.985  && 0.912 & 0.988 &  \textbf{0.989} & 0.986 \\ 
\midrule
\textbf{Fashion MNIST} &    0.734 & 0.870 &  \textbf{0.884} & 0.850  && 0.747 & 0.869 &  \textbf{0.884} & 0.851 \\ 
\midrule
\textbf{CIFAR-10}      & 0.568 & 0.885 &  0.122 & \textbf{0.888}  && 0.564 & \textbf{0.891} &  0.115 & 0.889 \\
\bottomrule
\end{tabular}}
    \caption{\textit{accuracy} and $accuracy_{10}$ comparing PoFL, KFC, and the baselines in \textbf{scenario B} under a Byzantine attack.}  
    \label{tab:kfcbyzantine_b}
\end{table*}

\begin{table*}[h!]
\resizebox{\linewidth}{!}{
\centering
\begin{tabular}{lrrlllrrll}
\toprule
 & \multicolumn{4}{c}{\textit{accuracy}} & & \multicolumn{4}{c}{\textit{$accuracy_{10}$}}\\
 \toprule
\textbf{}     & \textbf{PoFL}    & \textbf{KFC}  &  \textbf{Trim.-}&\textbf{Krum} && \textbf{PoFL}    & \textbf{KFC}  & \textbf{Trim.-}&\textbf{Krum}\\
 & & & \textbf{mean}&  && & & \textbf{mean}&\\ 
\midrule
\textbf{EMNIST}        &   0.992&  0.988&  \textbf{0.993}&0.985 && 0.992& 0.988& \textbf{0.993}&0.985\\ 
\midrule
\textbf{Fashion MNIST}&    0.905& 0.870&  \textbf{0.905}&0.853 &&  0.904&   0.868& \textbf{0.906}&0.853\\ 
\midrule
\textbf{CIFAR-10}      & \textbf{0.962}& 0.890&  0.158&0.886 && \textbf{0.960}& 0.887& 0.165&0.886\\
\bottomrule
\end{tabular}}
    \caption{\textit{accuracy} and $accuracy_{10}$ comparing \ecml{the baselines} and KFC in the \textbf{scenario A} under a Byzantine attack.}  \label{tab:kfcbyzantine_a}
\end{table*}

\begin{figure}[h!]
\begin{subfigure}{0.49\linewidth}
  \centering
  \includegraphics[width=\linewidth]{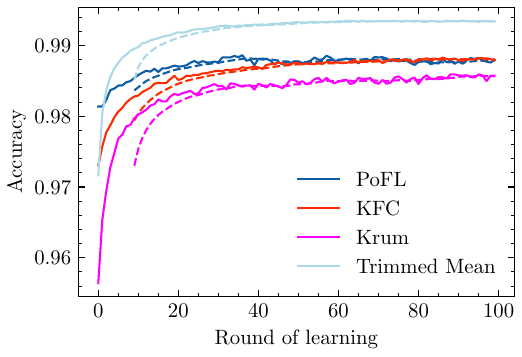}
  \caption{EMNIST (Original task).}
\end{subfigure} 
\begin{subfigure}{0.49\linewidth}
  \centering
  \includegraphics[width=\linewidth]{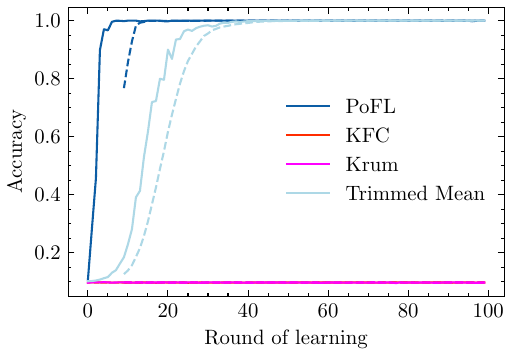}
  \caption{EMNIST (Backdoor task).}
\end{subfigure} 
\vskip\baselineskip
\begin{subfigure}{0.49\linewidth}
  \centering
  \includegraphics[width=\linewidth]{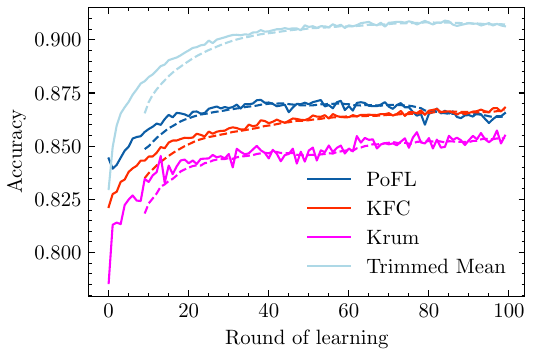}
  \caption{Fashion (Original task).}
\end{subfigure} 
\begin{subfigure}{0.49\linewidth}
  \centering
  \includegraphics[width=\linewidth]{kfc_vs_pofl_backdoor_fashion_all.pdf}
  \caption{Fashion (Backdoor task).}
\end{subfigure}
\vskip\baselineskip
\begin{subfigure}{0.49\linewidth}
  \centering
  \includegraphics[width=\linewidth]{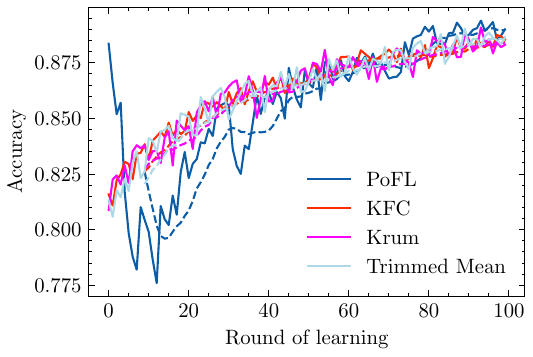}
  \caption{CIFAR (Original task).}
\end{subfigure}
\begin{subfigure}{0.49\linewidth}
  \centering
  \includegraphics[width=\linewidth]{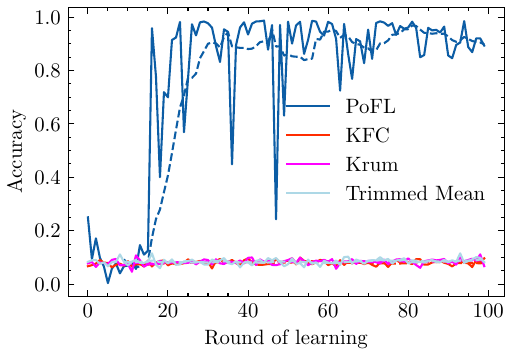}
  \caption{CIFAR (Backdoor task).}
\end{subfigure}
\vskip\baselineskip
\caption{$accuracy$ (line) and $accuracy_{10}$ (dashed line) of the original and backdoor tasks on EMNIST (a and b), Fashion MNIST (c and d) and CIFAR (e and f) datasets, respectively, in \textbf{scenario B}.}
\end{figure}

\begin{figure}[!ht]
\begin{subfigure}{0.49\linewidth}
  \centering
  \includegraphics[width=\linewidth]{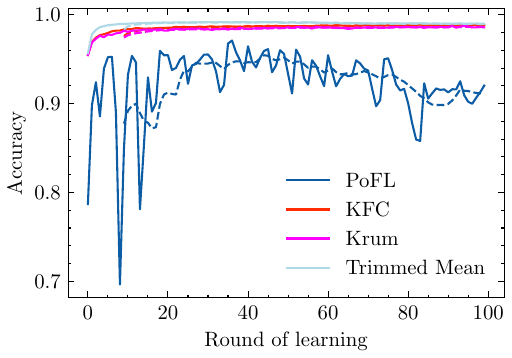}
  \caption{EMNIST.}
\end{subfigure} 
\begin{subfigure}{0.49\linewidth}
  \centering
  \includegraphics[width=\linewidth]{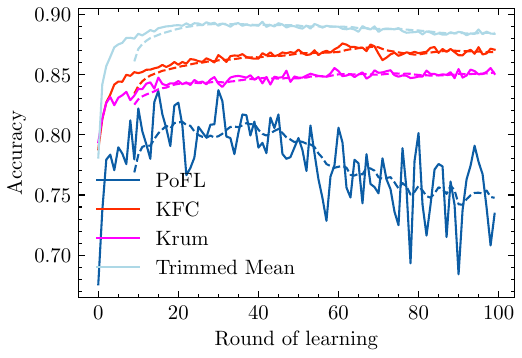}
  \caption{Fashion EMNIST.}
\end{subfigure} 
\vskip\baselineskip
\centering
\begin{subfigure}{0.49\linewidth}
  \centering
  \includegraphics[width=\linewidth]{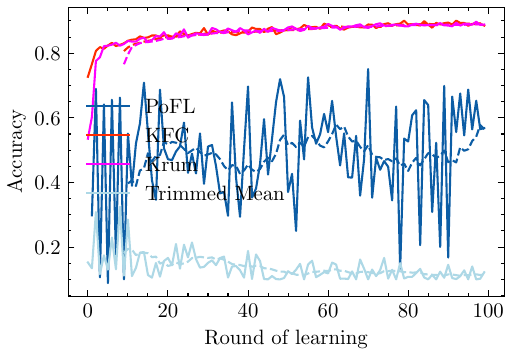}
  \caption{CIFAR.}
\end{subfigure}
\vskip\baselineskip
\caption{$accuracy$ (line) and $accuracy_{10}$ (dashed line) under the Byzantine attack on EMNIST (a), Fashion MNIST (b) and CIFAR (c) datasets, respectively, in \textbf{scenario B}.}
\end{figure}

\end{document}